\pdfoutput=1

\documentclass[11pt]{article}

\usepackage[final]{acl}

\usepackage{times}
\usepackage{latexsym}

\usepackage[T1]{fontenc}

\usepackage[utf8]{inputenc}

\usepackage{microtype}

\usepackage{inconsolata}

\usepackage{graphicx}

\usepackage{xcolor}
\usepackage{tcolorbox}

\usepackage{lipsum}
\usepackage{geometry}

\usepackage{microtype}
\usepackage{url}
\usepackage{booktabs}

\usepackage{subfigure}
\usepackage{enumitem}
\newenvironment{itemize*}%
 {\leftmargini=20pt\begin{itemize}%
  \setlength{\itemsep}{3pt}%
  \setlength{\parskip}{0pt}%
  }%
 {\end{itemize}}
\newenvironment{enumerate*}%
 {\begin{enumerate}%
  \setlength{\itemsep}{0pt}%
  \setlength{\parskip}{0pt}}%
 {\end{enumerate}}

\usepackage{amsmath}
\usepackage{cleveref}
\usepackage{listings}
\usepackage{titlesec}
\usepackage{multirow}
\usepackage{makecell}

\definecolor{lightred}{RGB}{255,163,163}
\definecolor{deepred}{RGB}{146,0,0}

\definecolor{midnightgreen}{rgb}{0.0, 0.29, 0.33}
\definecolor{deepgreen}{HTML}{0aa344}
\definecolor{deeppurple}{HTML}{7030a0}
\definecolor{deepblue}{HTML}{171d91}
\definecolor{brown}{HTML}{843c0c}
\definecolor{shadered}{HTML}{ffe5e5}
\definecolor{shadegreen}{HTML}{e5f7ed}

\newcommand{\red}{\textcolor{red}}
\newcommand{\green}{\textcolor{deepgreen}}

\NewDocumentCommand{\heng}
{ mO{} }{\textcolor{red}{\textsuperscript{\textit{Heng}}\textsf{\textbf{\small[#1]}}}}
\NewDocumentCommand{\cheng}
{ mO{} }{\textcolor{orange}{\textsuperscript{\textit{Cheng}}\textsf{\textbf{\small[#1]}}}}
\NewDocumentCommand{\yuji}
{ mO{} }{\textcolor{cyan}{\textsuperscript{\textit{Yuji}}\textsf{\textbf{\small[#1]}}}}

\title{\raisebox{-0.15cm}{\includegraphics[width=0.8cm]{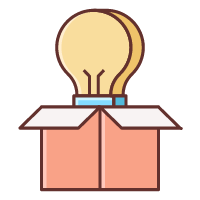}} \hspace{0.0cm} EscapeBench: Towards Advancing Creative Intelligence of \\ Language Model Agents}

\author{
Cheng Qian$^{1}$,  Peixuan Han$^{1}$, Qinyu Luo$^{2}$, Bingxiang He, Xiusi Chen$^{1}$, Yuji Zhang$^{1}$,\\
\textbf{Hongyi Du$^{1}$, Jiarui Yao$^{1}$, Xiaocheng Yang$^{1}$, Denghui Zhang$^{1,3}$, Yunzhu Li$^{4}$, Heng Ji$^{1}$}\\
$^{1}$University of Illinois Urbana-Champaign, $^{2}$Johns Hopkins University,\\
$^{3}$Stevens Institute of Technology, $^{4}$Columbia University\\
\texttt{\{chengq9, hengji\}@illinois.edu}\\
}

\begin{document}
\maketitle
\begin{abstract}
Language model agents excel in long-session planning and reasoning, but existing benchmarks primarily focus on goal-oriented tasks with explicit objectives, neglecting creative adaptation in unfamiliar environments. To address this, we introduce \textbf{EscapeBench}—a benchmark suite of room escape game environments designed to challenge agents with creative reasoning, unconventional tool use, and iterative problem-solving to uncover implicit goals.
Our results show that current LM models, despite employing working memory and Chain-of-Thought reasoning, achieve only 15\% average progress without hints, highlighting their limitations in creativity. To bridge this gap, we propose \textbf{EscapeAgent}, a framework designed to enhance creative reasoning through \textit{Foresight} (innovative tool use) and \textit{Reflection} (identifying unsolved tasks). Experiments show that EscapeAgent can execute action chains over 1,000 steps while maintaining logical coherence. It navigates and completes games with up to 40\% fewer steps and hints, performs robustly across difficulty levels, and achieves higher action success rates with more efficient and innovative puzzle-solving strategies.
All the data and codes are released\footnote{\url{https://github.com/qiancheng0/EscapeBench}}.
\end{abstract}

\section{Introduction}
Building robust language model (LM) agents to perform planning and reasoning has always been a challenging task.
Recent efforts have explored how agents could compress and utilize memory \citep{wang2023augmenting, hu2023chatdb, liu2023think, liang2023unleashing, wang2024jarvis, zhong2024memorybank}, perform complex reasoning \citep{wei2022chain, kojima2022large, zhou2023least, lin2024swiftsage, yao2023react}, planning \citep{wang2023describe, liu2023llm+, hao2023reasoning, yao2024tree, zhou2024language}, and reflection \citep{madaan2024self, zhang2024self, zhang2024agent, miao2024selfcheck, dhuliawala2024chain} to improve task success rate. Integrating these capabilities, recent lines of work begin to build agents for embodied actions \citep{zheng2024towards, huang2024embodied, zhu2023ghost} and tool use ~\citep{schick2023toolformer, qin2023tool, qian2024toolink, wang2025otc, qian2025toolrl} grounded in environments including the Web \citep{nakano2021webgpt, furuta2024multimodal, gur2024real}, games \citep{guo2023suspicion, xu2023exploring, hu2024survey}, and society \citep{park2023generative, liu2023training, li2023camel, ren2024emergence}.

\begin{figure}[!t]
    \centering
    \subfigure{\includegraphics[width=\linewidth]{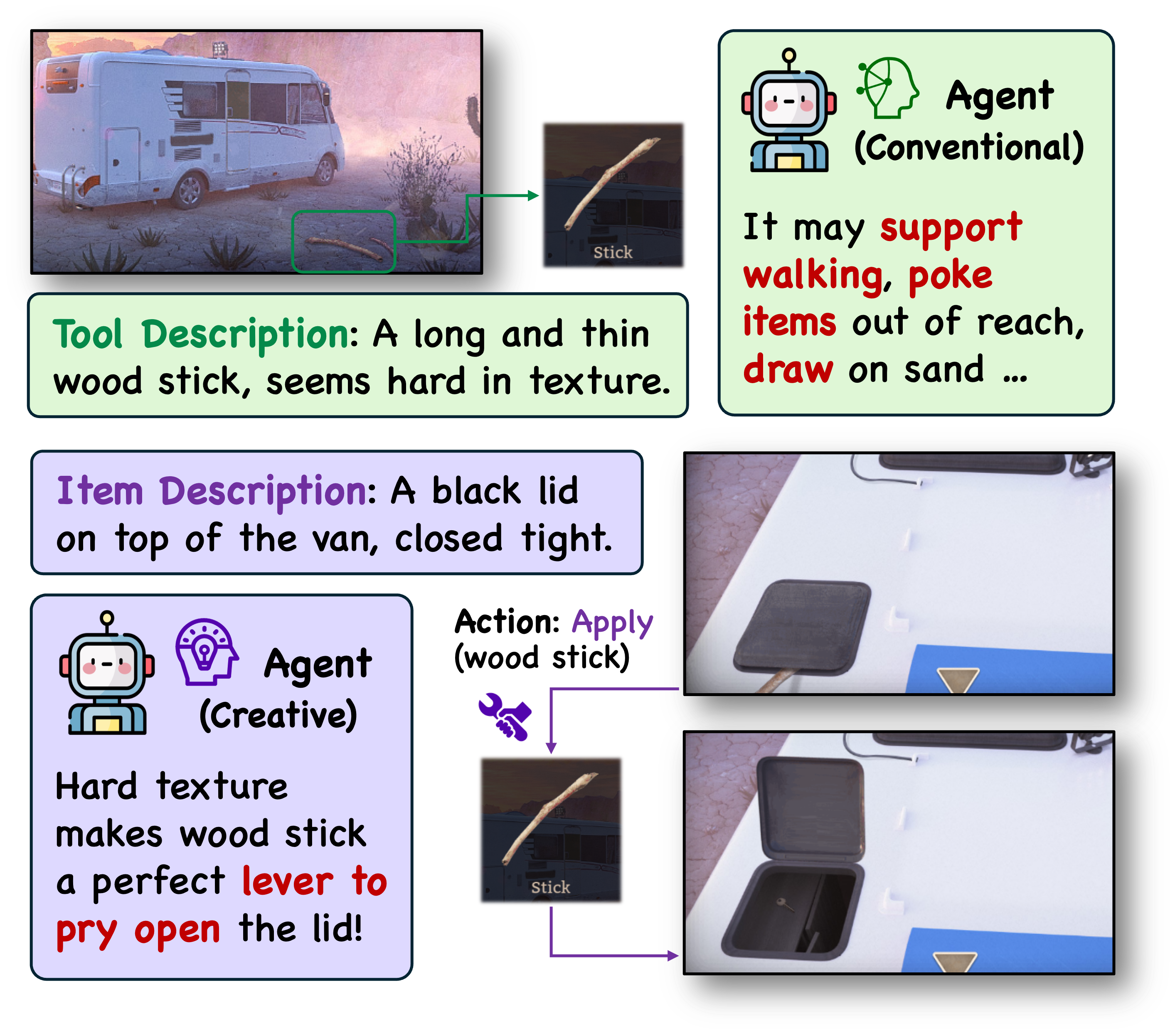}}
    \caption{An agent with creative thinking should adapt its observation (e.g. hard texture of wood stick) into a novel tool-use strategy (e.g. prying objects open).}
    \label{fig:intro}
\end{figure}

The surge of LM agent systems also accelerates the development of simulation environments, including tasks like computer-based operations~\citep{yao2022webshop, deng2024mind2web, zhou2024webarena, xie2024osworld, liu2024agentbench}, scientific research~\citep{wang2022scienceworld, bran2023chemcrow, boiko2023emergent, huang2023benchmarking}, and interactive experiences in text-based~\citep{cote2019textworld, urbanek2019learning, o2023hoodwinked, wu2024smartplay} or virtual sandbox game environments~\citep{lin2023agentsims, baai2023plan4mc, wang2024voyager}. However, most existing benchmarks are usually goal-oriented, emphasizing models' planning, reasoning, and error-handling abilities, while overlooking their \textit{creativity}: the capacity to think innovatively and adapt their observations to new, unstructured scenarios.

Current agents still significantly lack creativity in novel tool use~\citep{zhang2023creative}, as their training predominantly focuses on memorizing tool-task associations. This emphasis overshadows their ability to explore tool affordances and adapt to unstructured scenarios~\citep{zhang2024knowledge}. Despite this, creativity is still widely recognized as a crucial component of intelligence. In cognitive science, the well-established Triarchic Theory of Intelligence~\citep{sternberg1984toward} divides intelligence into three components: \textit{practical}, \textit{analytical}, and \textit{creative}. While current reasoning benchmarks primarily assess analytical intelligence through problem-solving, and simulation environments focus on practical intelligence by testing knowledge application in real-world scenarios, creative intelligence remains largely unaddressed.

To bridge this gap, we introduce \textbf{EscapeBench}, a benchmark that evaluates LM's creative reasoning using scenarios inspired by \textit{room escape} games. These scenarios challenge conventional thinking through unusual settings and require ``thinking outside the box'' skills including creative tool use and strategic problem-solving. As shown in \Cref{fig:intro}, a wooden stick, typically used for walking or poking, has to be repurposed to pry open a lid due to its hard texture. This demands agents to perform adaptive reasoning under customized constraints. Overall, our benchmark has several distinctive features:
\begin{itemize}[topsep=2pt, partopsep=-3pt, leftmargin=8pt, itemsep=-3pt]
\item \textbf{Creative Tool Use}: The tools at hand might be repurposed for creative use in order to solve the puzzle. These innovative ways of tool use are uncommon in the LM agent's existing parametric knowledge, requiring it to reason creatively and adapt its observation into customized scenarios.
\item \textbf{Uncertain Goal Pathways}: While the final goal of each game is escaping from the room, the pathways to achieving it cannot be explicitly foreseen. An agent cannot devise precise, long-range plans initially and must rely on trial and error to discover viable strategies.
\item \textbf{Super-Long Reasoning Chain}: Each scenario requires even an omniscient agent to perform over 100 steps, with at least 40 bottleneck actions required to achieve the goal. A human player may take up to an hour to complete one game.
\end{itemize}

We benchmarked multiple models within the BaseAgent framework, which incorporates working memory and Chain-of-Thought reasoning~\citep{wei2022chain}. Our results show that even the best models struggle to complete the easiest game setting without hints, often requiring up to ten times the optimal steps and falling far behind human performance. These findings highlight how models tend to be constrained by conventional thinking patterns, struggling to break free and show creativity.

To overcome this limitation, we introduce \textbf{EscapeAgent}, enhanced with \textit{Foresight} for creative tool use and \textit{Reflection} for implicit goal identification. Foresight enables the agent to propose and evaluate tool-use hypotheses before acting, while Reflection maintains an unsolved task list to guide future actions. Experiments show EscapeAgent reduces hint reliance by nearly 50\%, lowers total action steps, and performs robustly across difficulty levels, achieving more efficient progress and higher action success rates with creative strategies.
Our contributions include the following:
\begin{itemize}[topsep=2pt, partopsep=-5pt, leftmargin=8pt, itemsep=-4.5pt]
\item We identify challenges in LLM agent creative intelligence and introduce EscapeBench, a robust environment for evaluating agent creativity. 
\item We present EscapeAgent, which boosts creative reasoning by identifying implicit goals and generating innovative hypotheses.
\item We propose measuring creativity through tool use and crafting, and introduce new metrics that provide a fresh dimension for agent evaluation.
\end{itemize}


\section{Related Work}

\paragraph{Creativity in Language Models.}
Creativity is a cornerstone of human intelligence and a growing focus in AI research \cite{legg2007universal,lake2017building}. LMs have demonstrated notable creative capabilities across domains - they excel at generating narratives and poetry \citep{brown2020language, akoury2020storium}, show effectiveness in tool creation and design \citep{qian2023creator, cai2023large}, model real-world challenges~\citep{qian2025modelingagent}, and augment human creativity through interactive ideation \citep{mialon2023augmented}. In scientific discovery, research has also found that LM-generated ideas tend to be more novel but slightly less feasible than those from human experts \citep{si2024can,scimon2024}.

However, research on LM creativity still remains nascent, emphasizing novelty, surprise, and practical value through psychological assessments like the Alternative Uses Test (AUT)\citep{guilford1967creativity} and Torrance Tests of Creative Thinking (TTCT) \citep{boden1998creativity}. Creativity in LMs is categorized as combinatorial, exploratory, or transformational, with transformational being the most challenging \citep{franceschelli2023creativity}. A TTCT study found GPT-4 performing in the top 1\% of human fluency and originality, but adapting such assessments to other LMs faces limitations like sample randomness and high evaluation costs \citep{GUZIK2023100065}. Similarly, a modified Torrance Test \citep{zhao2024assessingunderstandingcreativitylarge} identified strengths in elaboration and originality but highlighted gaps influenced by prompts and role-play. Notably, most research evaluates backbone models, whereas our work explores creativity within an LM agent-based setting that requires complex reasoning and planning.

\paragraph{Agent Evaluation in Simulated Environment.}
Agent evaluation often focuses on text-based or sandbox environments for assessing cognitive and behavioral abilities in goal-oriented tasks \cite{zhou2023webarena, chen2022reliable, chen2024playbest, yu2024finconsynthesizedllmmultiagent, deng2024mind2web}, with emerging work exploring LM/VLM-enabled agents in robotics for real-world challenges \cite{liang2023code, huang2023voxposer, huang2024rekep, rana2023sayplan, zhu2025multiagentbench, yang2025embodiedbench}. Text-based environments \cite{yuan2018counting, cote2019textworld}, such as interactive fiction games \cite{lin2024swiftsage} or conversational agents \cite{GameEval}, evaluate natural language understanding, reasoning, and decision-making consistency \cite{uludaugli2023non,qi2024civrealm}. Games like Zork \cite{zork1980} and TextWorld measure narrative comprehension and problem-solving in structured contexts. In contrast, sandbox environments \cite{lin2023agentsims, gan2021threedworld, fan2022minedojo} like Minecraft \cite{zhu2023ghost} and Roblox \cite{rospigliosi2022metaverse} provide more open-ended settings that test spatial reasoning, planning, and collaboration \cite{carroll2019utility, multi_agent_coordination}. The settings typically rely on task-specific metrics for goal achievement but overlook creative and proactive problem-solving in unfamiliar contexts. To address this, we introduce EscapeBench to evaluate agents' creative reasoning in navigating uncertain goal pathways, offering a novel approach to agent assessment.
\begin{figure*}[!t]
    \centering
    \subfigure{\includegraphics[width=0.9\linewidth]{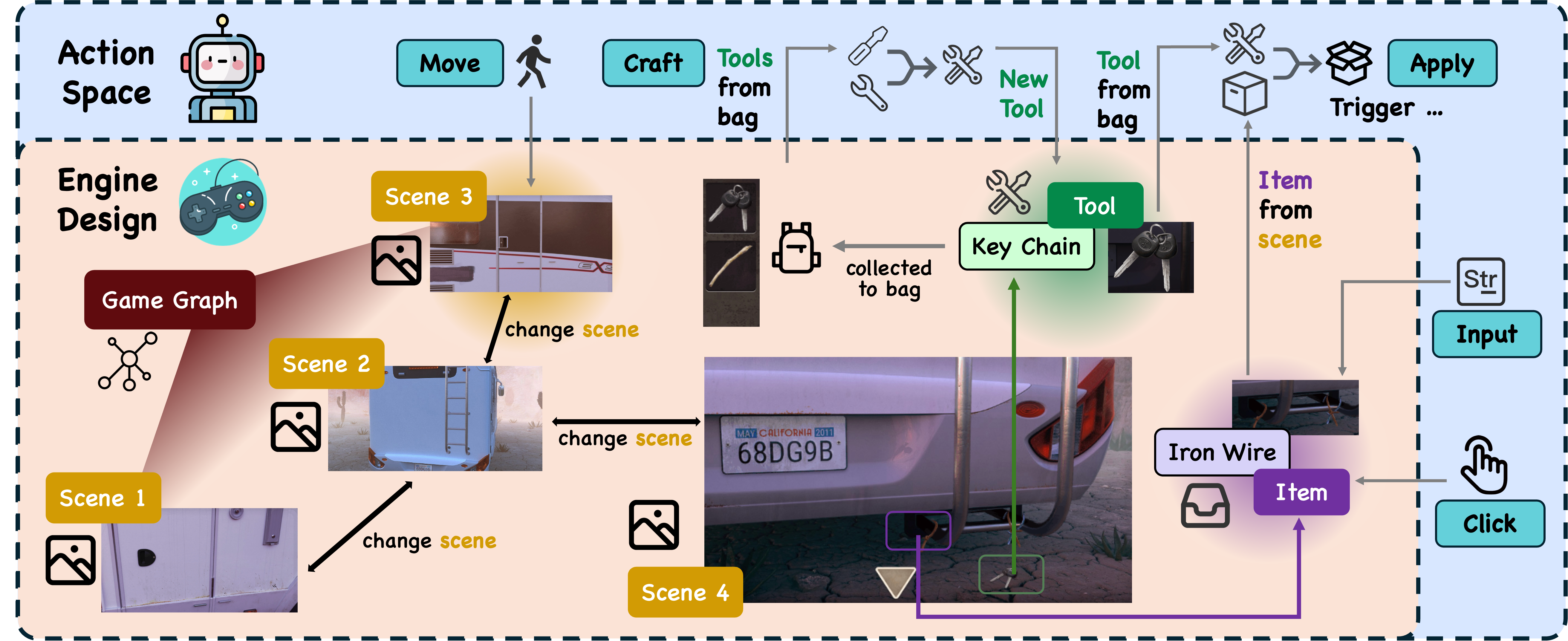}}
    \caption{An illustration of \textcolor[HTML]{D29B03}{Scenes}, \textcolor[HTML]{03894A}{Tools}, and \textcolor[HTML]{70309F}{Items} in the game and their relations with agent action space. Tools can be collected for ``Apply'' and ``Craft'', while items require ``Input'', ``Click'' or ``Apply'' of tools to trigger effects.}
    \label{fig:engine_action}
\end{figure*}

\section{EscapeBench Construction}
\label{sec:in3_dataset}
Most agent benchmarks focus on explicit, goal-oriented tasks grounded in commonsense knowledge, where agents can chart clear pathways to achieve goals using analytical and practical intelligence, but they often overlook creative intelligence. This raises our core research question: \textbf{How to build an environment that benchmarks an agent's creative intelligence?} Given that tool use is central to agent functionality, we propose room escape game scenarios, which naturally require creative tool use to solve complex puzzles, as an ideal environment for this evaluation.

\subsection{Engine Design}
\label{sec:engine_design}
Our game engine aims to simulate the room escape environment that \textbf{i)} receives agent actions and \textbf{ii)} makes corresponding environment feedback as agent action's reward. Specifically, our game engine involves three key components:
\begin{itemize}[topsep=2pt, partopsep=-5pt, leftmargin=8pt, itemsep=-4.5pt]
\item \textbf{\textcolor[HTML]{D29B03}{Scenes}}: The container of tools and items, connected with each other forming a graph structure that constitutes the whole game scenario.
\item \textbf{\textcolor[HTML]{70309F}{Items}}: Objects that are intractable in each scene. Tools, inputs, and other interactions may be applied to trigger its state change or other effects.
\item \textbf{\textcolor[HTML]{03894A}{Tools}}: Objects that could be collected in each scene, usually applied to other items to take effect or to other tools to craft new ones.
\end{itemize}
The interaction of these components defines a game's basic logic. In the van example from \Cref{fig:engine_action}, scenes are connected into a graph, representing physical connectivity via doors or tunnels. The tool \textit{key chain} is collected in the bag for future use, while the \textit{wire iron} awaits something sharp to cut it to trigger effects. Please refer to \Cref{apdx:engine_design_details} for more detailed examples and explanations.

\subsection{Action Space}
\label{sec:action_space}
The model agent could take five different actions. While the action space is well-defined, the parameter space—regarding the scenes, items, or tools involved in these actions—is high-dimensional, thus allowing for dynamic interactions.
\begin{itemize}[topsep=2pt, partopsep=-5pt, leftmargin=8pt, itemsep=-4.5pt]
\item \textbf{Move (Scene)}: Move to an adjacent scene.
\item \textbf{Click (Item)}: Click to simply interact with an item in the scene.
\item \textbf{Apply (Tool, Item)}: Apply a tool in the bag to an item in the scene.
\item \textbf{Input (\textit{str}, Item)}: Input an arbitrary string to an item in the scene.
\item \textbf{Craft (Tool, Tool)}: Use two tools in the bag to craft a new one.
\end{itemize}
\Cref{fig:engine_action} illustrates the connections between game engine components and agent action space. Among all the actions, ``Apply'' and ``Craft'' stand out as the most creativity-driven, as they require the agent to think innovatively about how to use or craft tools in an unseen way during its training. We delve into specific examples in \Cref{sec:preliminary_study}.

\begin{figure}[!t]
    \centering
    \subfigure{\includegraphics[width=0.96\linewidth]{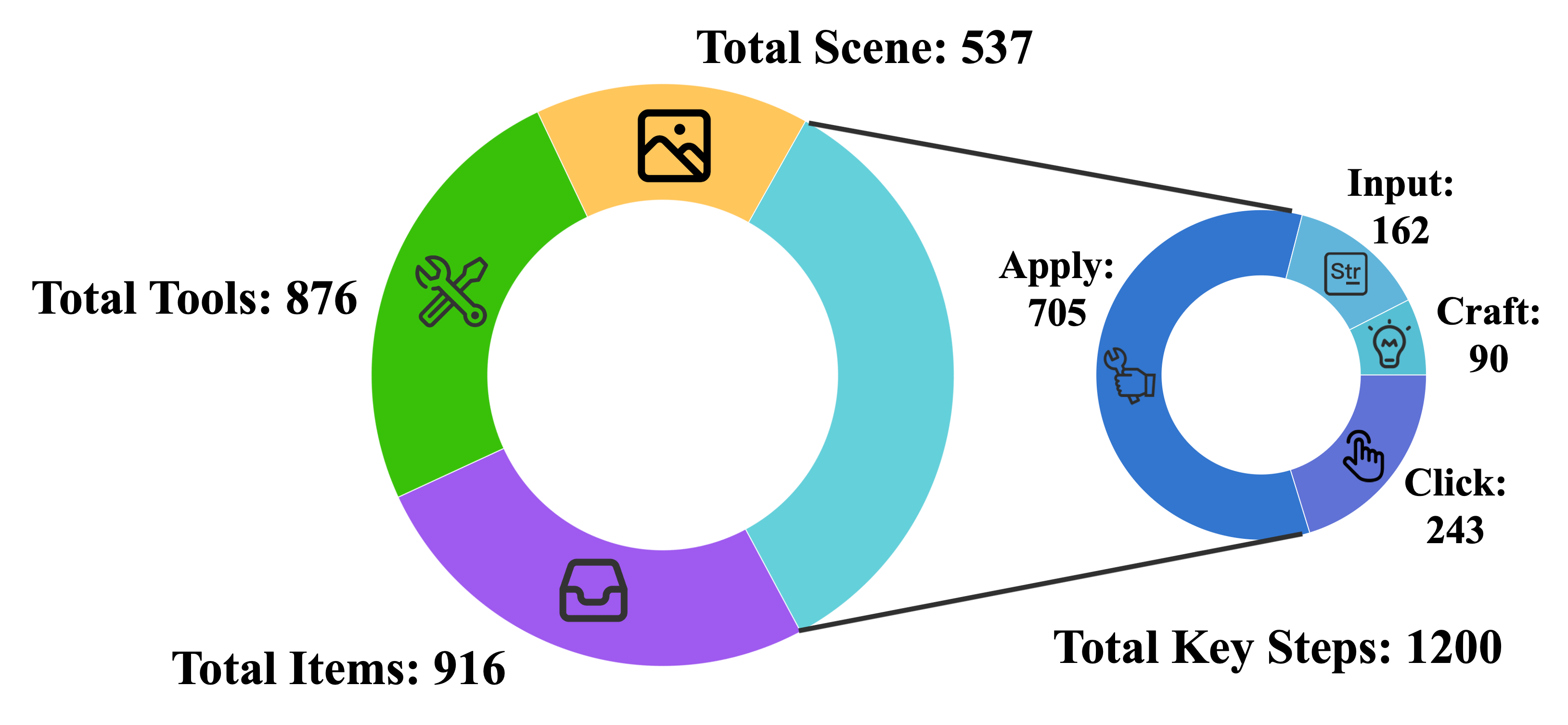}}
    \caption{Statistics of total \textcolor[HTML]{D29B03}{Scenes}, \textcolor[HTML]{03894A}{Tools}, and \textcolor[HTML]{70309F}{Items} across all game settings. ``Key Steps'' refer to the essential bottleneck actions required to complete the game.}
    \label{tab:game_statistic}
\end{figure}

\begin{table*}[!t]
    \centering
    \subfigure{\includegraphics[width=0.95\linewidth]{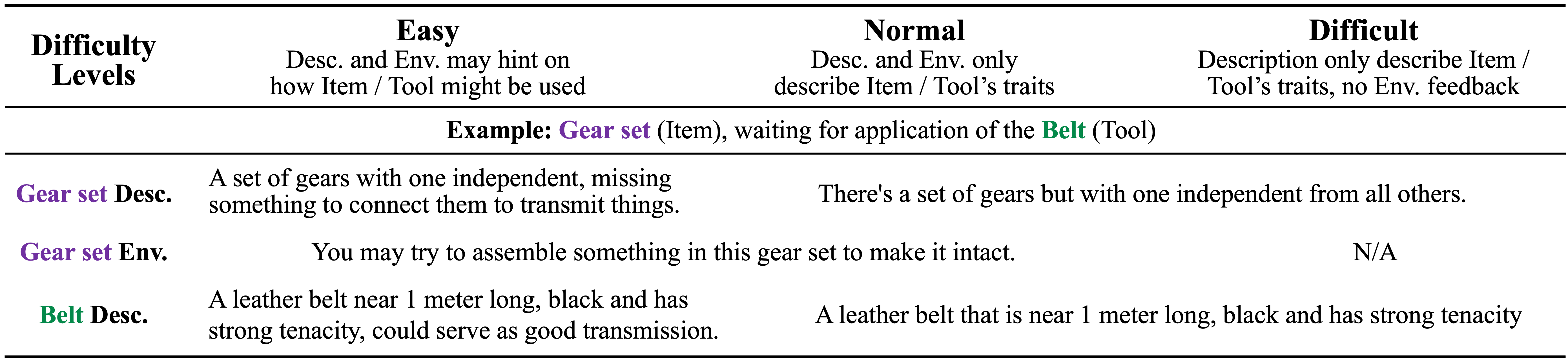}}
    \vspace{-3mm}
    \caption{Rules and examples for different difficulty levels. \textit{Desc.} refers to Item or Tool descriptions, while \textit{Env.} represents the game engine's feedback when an unexpected action targets an Item or Tool.}
    \label{tab:difficulty_division}
\end{table*}

\subsection{Annotation and Statistics}
\label{sec:annotation}
Building on existing online room escape games and puzzle-solving logic\footnote{\url{https://spotlight.ee}}, we introduce EscapeBench, featuring 36 game settings spanning three difficulty levels. Each scenario includes three versions with consistent solution logic but varying in description clarity and feedback granularity. A detailed example is shown in \Cref{tab:difficulty_division}.
All scenes, items, and tools are manually annotated to ensure high quality.
These scenarios emphasize creative tool use and crafting strategies, challenging agents throughout the game, making EscapeBench a robust environment for testing creativity. A detailed statistic is presented in \Cref{tab:game_statistic}.

\begin{table*}[!t]
    \centering
    \subfigure{\includegraphics[width=0.96\linewidth]{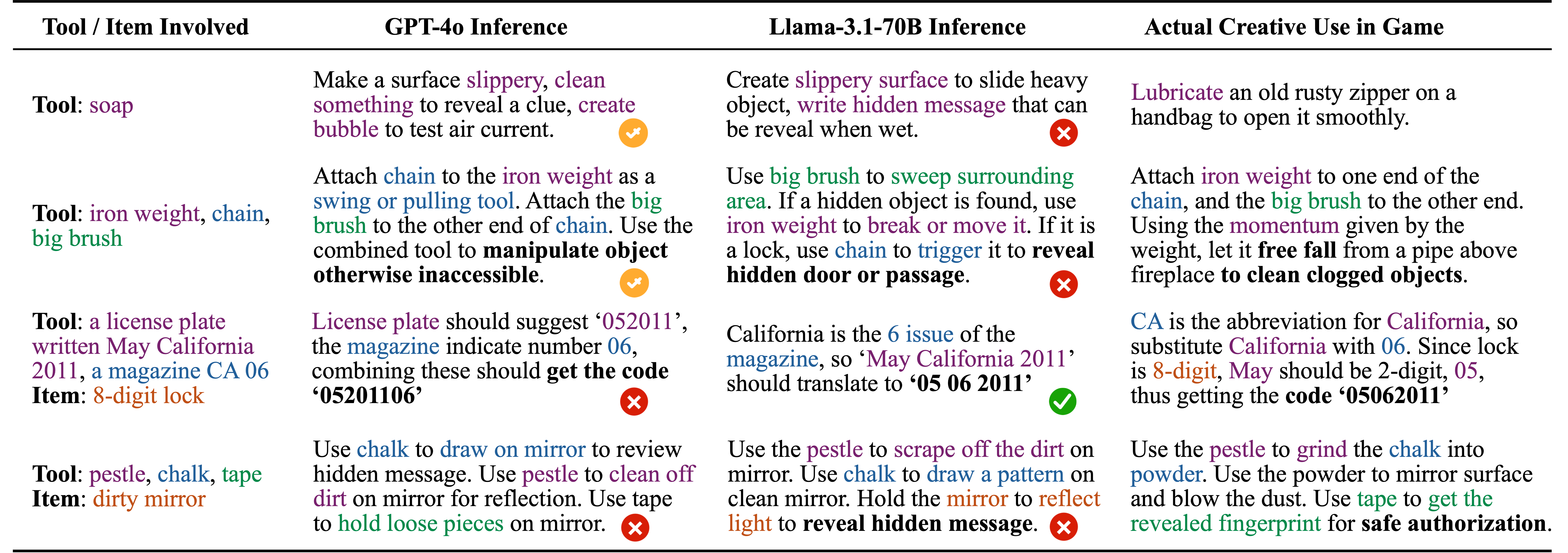}}
    \vspace{-3mm}
    \caption{Creative reasoning cases in EscapeBench and model's responses versus actual creative uses of tools.}
    \label{tab:preliminary_cases}
\end{table*}

\subsection{Preliminary Study}
\label{sec:preliminary_study}
We sample scenarios from EscapeBench and test GPT-4o~\citep{hurst2024gpt} and LLama-3.1-70B~\citep{dubey2024llama}'s creative reasoning performance through case studies in \Cref{tab:preliminary_cases}. Our results reveal that:
\textbf{i)} EscapeBench presents diverse creative reasoning challenges, including unconventional tool use, implicit numerical puzzles, and innovative tool crafting.
\textbf{ii)} Both closed- and open-source models struggle with creativity, especially in identifying implicit goals and forming creative strategies.
These findings highlight the complexity of EscapeBench and the gaps in model creativity.

\begin{figure*}[!t]
    \centering
    \subfigure{\includegraphics[width=0.96\linewidth]{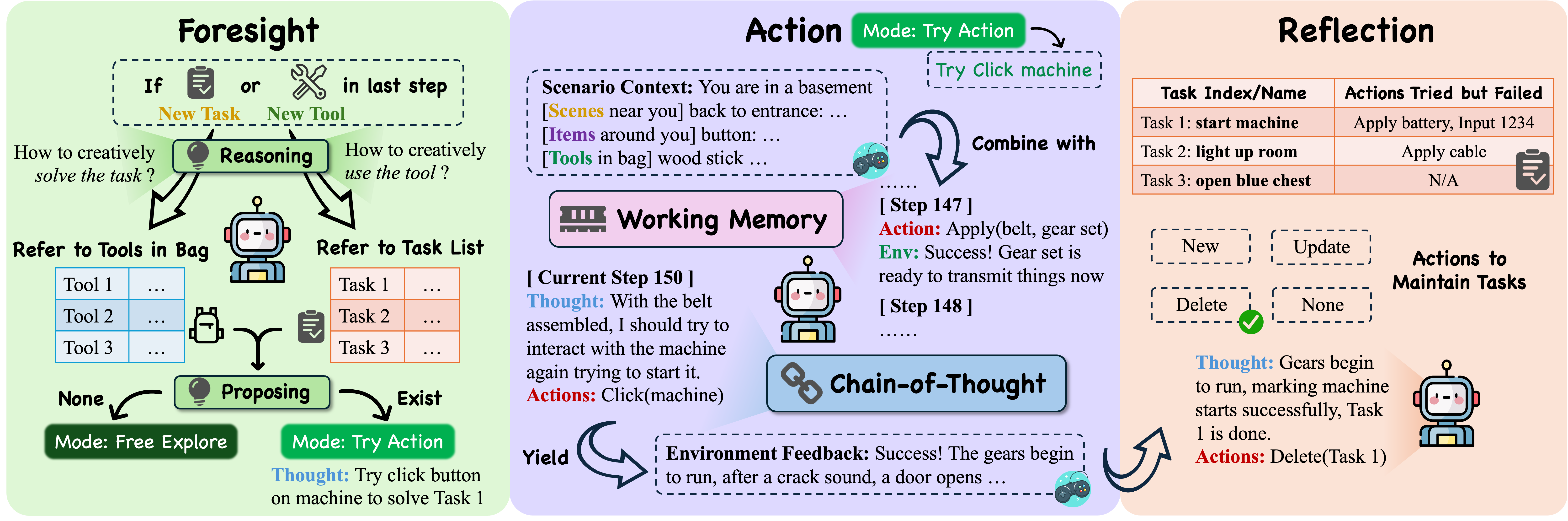}}
     \vspace{-3mm}
    \caption{Illustration of the EscapeAgent design. Building on the BaseAgent (\textit{Action}), we integrate the \textit{Foresight} and \textit{Reflection} modules to enhance the agent's capabilities in creative reasoning and implicit goal identification.}
    \label{fig:agent_design}
\end{figure*}

\section{EscapeAgent Design}
To address challenges identified in the preliminary study, we introduce \textbf{EscapeAgent}, a framework addressing two core issues from EscapeBench:
\begin{itemize}[topsep=2pt, partopsep=-3pt, leftmargin=8pt, itemsep=-3pt]
\item \textit{Uncertain Goal Pathways}: A \textbf{Reflection} module dynamically manages a task list, refining goals through trial-and-error to enhance action focus and proactive task discovery (\Cref{sec:reflection_module}).
\item \textit{Creative Tool Use}: A \textbf{Foresight} module enabling explicit reasoning about tool applications, allowing the agent to hypothesize and evaluate strategies before execution (\Cref{sec:foresight_module}).
\end{itemize}
Integrating both modules with a BaseAgent, EscapeAgent excels in handling \textit{Super-long Reasoning Chains} and significantly boosts the model's creativity, problem-solving, and strategic thinking.

\subsection{BaseAgent}
\label{sec:base_agent}
The \textbf{BaseAgent} serves as the foundation of EscapeAgent, as shown in the middle section of \Cref{fig:agent_design}. It takes actions based on the scenario context provided by the game engine and updates its working memory with environment feedback after each step. The working memory stores the agent’s previous actions and corresponding feedback.  

The BaseAgent employs Chain-of-Thought~\cite{wei2022chain} reasoning to decide its next action. Note that it serves as a strong baseline and standard evaluation method for EscapeBench, as it already incorporates all the essential components an agent needs for reasoning. For more implementation details, please refer to \Cref{apdx:base_agent}.


\subsection{Reflection Module}
\label{sec:reflection_module}
The \textbf{Reflection Module} manages a structured task list updated through three actions:
\begin{itemize}[topsep=2pt, partopsep=-5pt, leftmargin=8pt, itemsep=-4pt]
\item \textit{New}: Add a newly identified, unsolved task. 
\item \textit{Update}: Record attempted but failed actions. \item \textit{Delete}: Remove a task when its goal is achieved.
\end{itemize}
Each entry includes the task name, target item, and failed actions, preventing repeated mistakes and improving efficiency. Triggered after non-\textit{move} actions, it uses feedback to update the task list. For example, in \Cref{fig:agent_design} (right), Task 1 is deleted once the machine starts, encouraging focused problem-solving over random exploration. See \Cref{apdx:reflection_module} for details.

\subsection{Foresight Module}
\label{sec:foresight_module}
The \textbf{Foresight Module} enhances creative reasoning by explicitly evaluating tool use and problem-solving strategies. It activates in two cases:
\begin{itemize}[topsep=2pt, partopsep=-5pt, leftmargin=8pt, itemsep=-4pt]
\item \textit{New Task Identified}: The agent hypothesizes potential actions to achieve it using available tools.
\item \textit{New Tool Collected}: The agent assesses its use for solving existing tasks or crafting new tools.
\end{itemize}
If a valid hypothesis is proposed, the agent enters ``Try Action'' state to test it; otherwise, it stays in ``Free Explore'' state, operating like the BaseAgent. For example in \Cref{fig:agent_design} (left), the agent identifies clicking the button as an action worth trying, thus guiding it into the ``Try Action'' state. It enables the agent to adapt flexibly under customized scenarios, make bold hypotheses, and execute targeted trials efficiently. See \Cref{apdx:reflection_module} for details.

\begin{table*}[!t]
\begin{center}
\small
\tabcolsep=0.03\linewidth
\resizebox{1.0\linewidth}{!}{
\begin{tabular}{lccccc}
\toprule
\textbf{Model Name} & \textbf{$^{\downarrow}$Hints Used} & \textbf{$^{\downarrow}$Total Steps} & \textbf{\makecell{$^{\uparrow}$Early Exit\\ \ Progress (\%)}} & \textbf{\makecell{$^{\downarrow}$Tool Hints Used\\(percentage)}} & \textbf{\makecell{$^{\downarrow}$Key Steps Hints Used\\(percentage)}} \\
\midrule
GPT-4o  & 10.30 & 723.61 & 24.75 & 2.17 (8.55\%) & 8.14 (24.27\%) \\
GPT-4o-mini  & 15.19 & 1002.39 & 16.06 & 2.00 (8.97\%) & 13.19 (38.84\%) \\
Claude-3.5-Sonnet  & 8.97 & 690.31 & 28.95 & 1.34 (5.13\%) & 7.64 (22.44\%) \\
Gemini-1.5-pro  & 11.06 & 824.31 & 24.18 & 2.50 (9.89\%) & 8.56 (24.83\%) \\
\midrule
Llama-3.1-70B  & 14.53 & 982.42 & 19.00 & 3.11 (12.22\%) & 11.42 (33.29\%) \\
Qwen2.5-72B  & 16.50 & 1102.50 & 12.46 & 5.33 (20.97\%) & 11.17 (32.02\%) \\
DeepSeek-LLM-67B  & 25.50 & 1558.47 & 6.63 & 10.50 (42.95\%) & 15.00 (43.73\%) \\
Yi-1.5-34B  & 24.00 & 1573.33 & 11.96 & 8.11 (33.83\%) & 15.92 (46.18\%) \\
Phi-3-medium-128k  & 32.19 & 1871.19 & 7.34 & 12.11 (49.45\%) & 20.11 (59.1\%) \\
Llama-3.1-8B  & 25.86 & 1543.30 & 10.10 & 6.81 (28.56\%) & 19.11 (56.00\%) \\
Ministral-8B  & 25.31 & 1556.97 & 8.97 & 7.17 (29.90\%) & 18.19 (53.87\%) \\
Qwen2.5-7B  & 32.20 & 1950.42 & 6.52 & 13.81 (55.96\%) & 18.47 (54.43\%) \\
\midrule
Average Human  & 4.33 & 257.83 & 59.65 & 0.17 (0.69\%) & 4.17 (12.28\%) \\
\bottomrule
\end{tabular}
}
\end{center}
\vspace{-3mm}
\caption{Benchmarking results of \textbf{BaseAgent} with different core models on EscapeBench. An oracle action chain's total step is only 107.83 on average. Both closed- and open-source models rely heavily on hints to complete the escape compared to human performance, with smaller-scale models exhibiting a particularly high dependency.}
\label{tab:benchmark_results}
\end{table*}

\section{Experiments}
We divide experiments into: \textbf{i)} Benchmarking model creativity within the BaseAgent, and \textbf{ii)} Evaluating EscapeAgent's effectiveness.

\subsection{Settings}
\label{sec:exp_settings}
\paragraph{Environment.} Experiments are conducted on 36 game settings. An agent is considered to be \textbf{making progress} if it either \textbf{achieves a key step} (defined in \Cref{tab:game_statistic}) or \textbf{collects a new tool}. Agents will receive help if they fail to make progress for 50 consecutive actions (see \Cref{apdx:experiment_setting}), thus ensuring full completion of the game. The working memory length is set to 10.

\paragraph{Models.} We evaluate both closed- and open-source models: GPT-4o, GPT-4o-mini~\citep{hurst2024gpt}, Claude-3.5~\citep{2024claude}, Gemini-1.5~\citep{team2024gemini}, Llama-3.1~\citep{dubey2024llama}, Qwen-2.5~\citep{2024qwen2.5}, DeepSeek-LLM~\citep{liu2024deepseek}, Phi-3.5~\citep{abdin2024phi}, Yi~\citep{young2024yi}, Ministral~\citep{2024ministral}. Models with fewer than 7B parameters are excluded due to near-random behavior. For consistency, we set sampling temperature to $T=0$ and $n=1$.

\paragraph{Metrics.} We use two main metrics including:
\begin{itemize}[topsep=2pt, partopsep=-3pt, leftmargin=8pt, itemsep=-3pt]
\item \textbf{Hints Used}: Total hints used in a game.
\item \textbf{Total Steps}: Total actions taken in a game.
\end{itemize}
Auxiliary metrics for analysis include:
\begin{itemize}[topsep=2pt, partopsep=-3pt, leftmargin=8pt, itemsep=-3pt]
\item \textbf{Early Exit Progress}: Proportion of key steps and tools collected before needing a hint (game progress before needing a hint for the first time).
\item \textbf{Tool Hints Used (percentage)}: Hints used for tool collection (normalized by total tools).
\item \textbf{Key Step Hints Used (percentage)}: Hints used for key steps (normalized by total key steps).
\end{itemize} Results are micro-averaged across the 36 settings.

\begin{figure}[!t]
    \centering
    \subfigure{\includegraphics[width=0.9\linewidth]{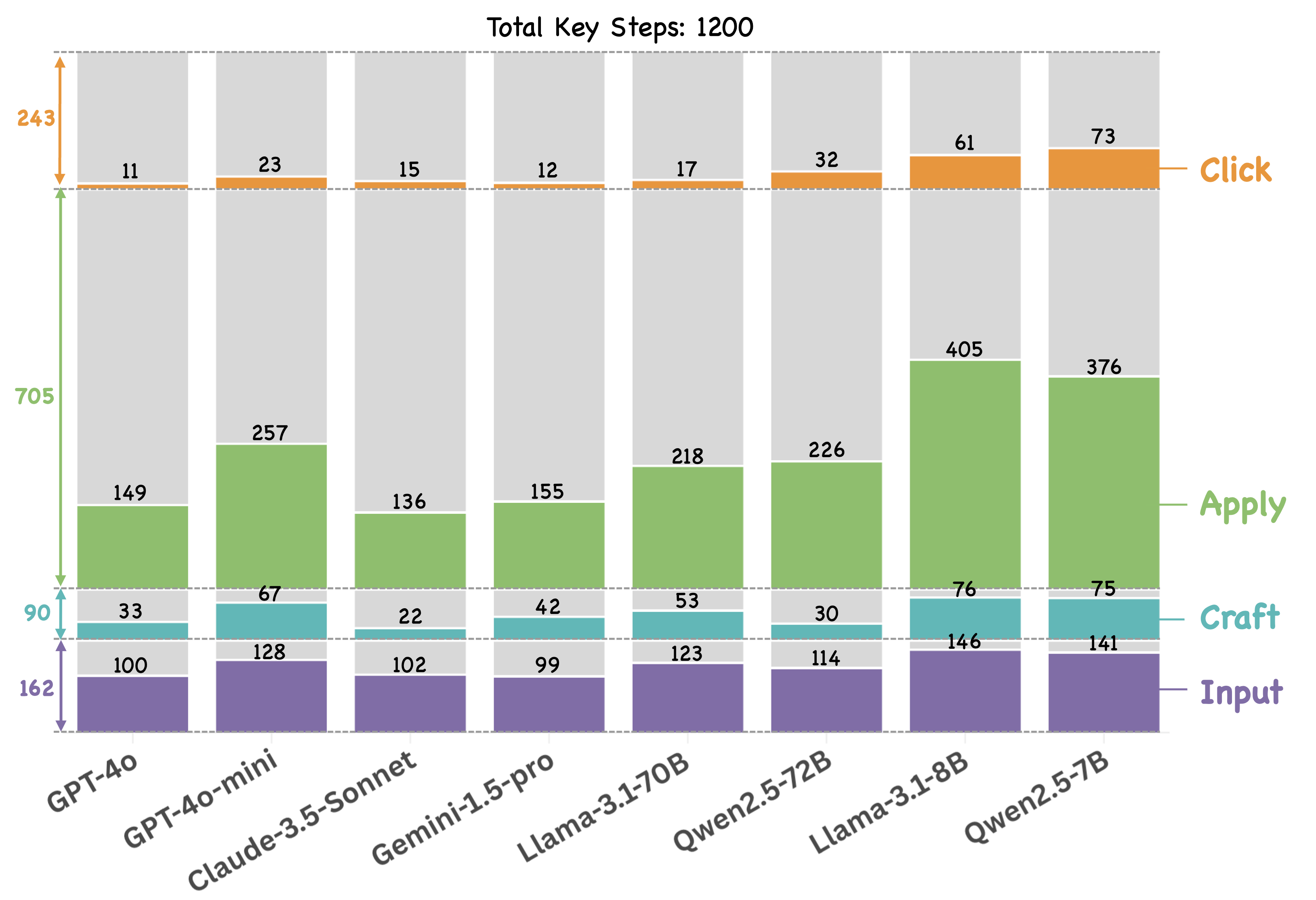}}
    \vspace{-3mm}
    \caption{Distribution of Key Steps Hints Used, categorized by different actions. Colored bars represent the percentage of hints used for each action type relative to the total key steps for that type (See right of \Cref{tab:game_statistic}).}
    \label{fig:help_distribution}
    \vspace{-3mm}
\end{figure}

\begin{table*}[!t]
    \centering
    \subfigure{\includegraphics[width=\linewidth]{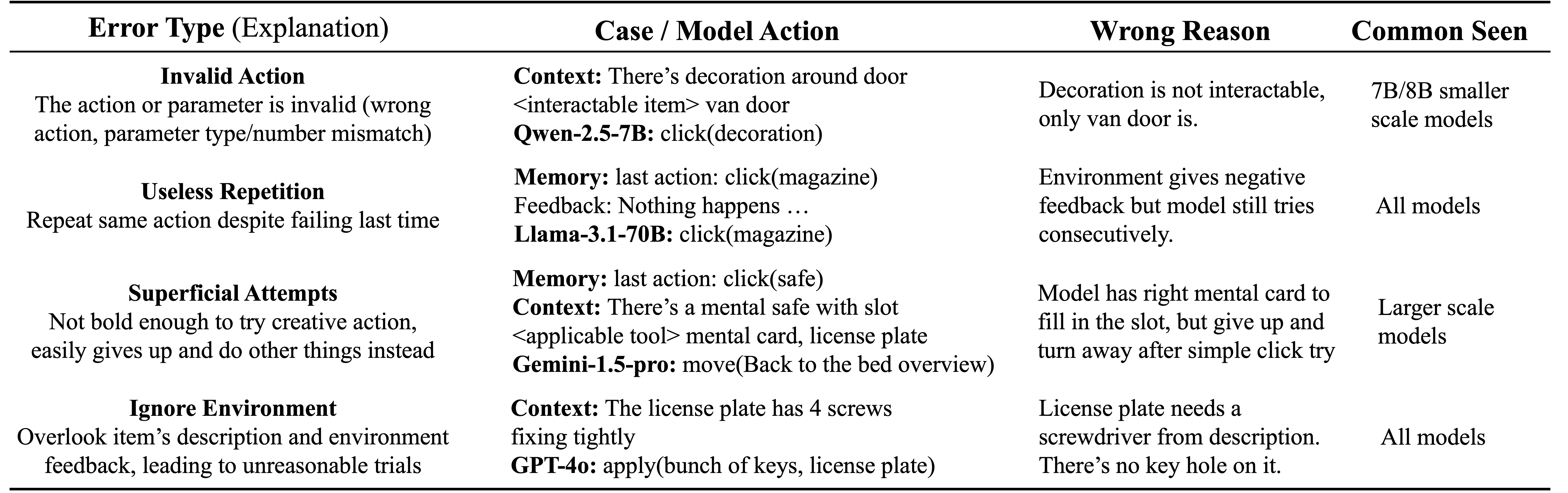}}
    \vspace{-3mm}
    \caption{Error Analysis of \textbf{BaseAgent}'s inefficiency or failures. All cases are selected in the same game setting.}
    \label{tab:error_analysis}
\end{table*}

\begin{table*}[!t]
\begin{center}
\small
\tabcolsep=0.03\linewidth
\resizebox{1.0\linewidth}{!}{
\begin{tabular}{lccccc}
\toprule
\textbf{Model Name} & \textbf{$^{\downarrow}$Hints Used} & \textbf{$^{\downarrow}$Total Steps} & \textbf{\makecell{$^{\uparrow}$Early Exit\\ \ Progress (\%)}} & \textbf{\makecell{$^{\downarrow}$Tool Hints Used\\(percentage)}} & \textbf{\makecell{$^{\downarrow}$Key Steps Hints Used\\(percentage)}} \\
\midrule
GPT-4o & $5.03_{\green{\downarrow 5.27}}$ & $452.75_{\green{\downarrow 270.86}}$ & $47.03_{\red{\uparrow 22.28}}$ & $0.33_{\green{\downarrow 1.84}}$ (1.19\%) & $4.70_{\green{\downarrow 3.44}}$ (13.74\%) \\
GPT-4o-mini & $10.58_{\green{\downarrow 4.61}}$ & $752.25_{\green{\downarrow 250.14}}$ & $28.17_{\red{\uparrow 12.11}}$ & $1.14_{\green{\downarrow 0.86}}$ (4.16\%) & $9.44_{\green{\downarrow 3.75}}$ (28.53\%) \\
\midrule
Llama-3.1-70B-Instruct & $7.92_{\green{\downarrow 6.61}}$ & $645.19_{\green{\downarrow 337.23}}$ & $31.44_{\red{\uparrow 12.44}}$ & $1.42_{\green{\downarrow 1.69}}$ (5.44\%) & $6.56_{\green{\downarrow 4.86}}$ (19.15\%) \\
Qwen2.5-72B-Instruct & $9.72_{\green{\downarrow 6.78}}$ & $746.61_{\green{\downarrow 355.89}}$ & $28.62_{\red{\uparrow 16.16}}$ & $2.72_{\green{\downarrow 2.61}}$ (10.61\%) & $7.00_{\green{\downarrow 4.17}}$ (20.71\%) \\
DeepSeek-LLM-67b-Chat & $20.14_{\green{\downarrow 5.36}}$ & $1285.03_{\green{\downarrow 273.44}}$ & $15.30_{\red{\uparrow 8.67}}$ & $10.81_{\red{\uparrow 0.31}}$ (42.72\%) & $9.34_{\green{\downarrow 5.66}}$ (27.20\%) \\
Yi-1.5-34B-Chat  & $22.59_{\green{\downarrow 1.41}}$ & $1468.03_{\green{\downarrow 105.30}}$ & $12.04_{\red{\uparrow 0.08}}$ & $10.42_{\red{\uparrow 2.31}}$ (41.61\%) & $12.19_{\green{\downarrow 3.73}}$ (35.71\%) \\
Phi-3-medium-128k-instruct  & $25.75_{\green{\downarrow 6.44}}$ & $1513.69_{\green{\downarrow 357.50}}$ & $9.99_{\red{\uparrow 2.65}}$ & $11.20_{\green{\downarrow 0.91}}$ (44.99\%) & $14.55_{\green{\downarrow 5.56}}$ (43.35\%) \\
Llama-3.1-8B-Instruct & $19.81_{\green{\downarrow 6.05}}$ & $1271.53_{\green{\downarrow 271.77}}$ & $19.22_{\red{\uparrow 9.12}}$ & $6.64_{\green{\downarrow 0.17}}$ (25.23\%) & $13.22_{\green{\downarrow 5.89}}$ (39.34\%) \\
Ministral-8B-Instruct & $19.47_{\green{\downarrow 5.84}}$ & $1233.72_{\green{\downarrow 323.25}}$ & $19.34_{\red{\uparrow 10.37}}$ & $6.61_{\green{\downarrow 0.56}}$ (25.56\%) & $12.86_{\green{\downarrow 5.33}}$ (38.93\%) \\
Qwen2.5-7B-Instruct & $27.53_{\green{\downarrow 4.67}}$ & $1639.58_{\green{\downarrow 310.84}}$ & $8.56_{\red{\uparrow 2.04}}$ & $13.00_{\green{\downarrow 0.81}}$ (52.66\%) & $14.58_{\green{\downarrow 3.89}}$ (43.99\%) \\
\bottomrule
\end{tabular}
}
\end{center}
\vspace{-3mm}
\caption{Benchmarking Results of \textbf{EscapeAgent} with different core models. Nearly all the performance raises compared to \Cref{tab:benchmark_results}, showcasing the effectiveness of EscapAgent in promoting the agent's creativity.}
\label{tab:creative_results}
\end{table*}

\subsection{Benchmarking Results}
\label{sec:base_agent_exp}
We benchmark current models using the BaseAgent framework, with results in \Cref{tab:benchmark_results} showing that large-scale closed-source models consistently outperform smaller models. Key insights include:
\begin{itemize}[topsep=2pt, partopsep=-3pt, leftmargin=8pt, itemsep=-3pt]
\item Most hints are used on key steps, which demand creative reasoning, while models may often collect tools through random exploration.
\item Models require significantly more action steps and hints than the average human, and up to \~ 20x more steps than the most efficient action chain.
\end{itemize}
We further present an ablation study of total hints and steps used in \Cref{apdx:ablation}.

\paragraph{Input and craft are the most challenging actions.} As shown in \Cref{fig:help_distribution}, while ``Apply'' actions require the most hints in absolute terms, ``Input'' and ``Craft'' actions have the highest relative hint usage compared to the total number of key steps. This likely reflects the large parametric space of ``Input'' actions, where random guesses are impractical, and the creativity-demanding nature of ``Craft'' actions.

\paragraph{Error analysis.} We observe from \Cref{tab:error_analysis} that the BaseAgent often gets stuck and relies on hints due to its struggles with environment-following and creativity. Smaller models tend to perform invalid actions in complex scenarios, while larger models excel at tool collection but fail to attempt creative actions, resorting to superficial strategies. Our EscapeAgent addresses this by promoting more purposeful and creative actions.

\subsection{EscapeAgent Results}
\label{sec:creative_agent_exp}
The introduction of the Reflection and Foresight in EscapeAgent, as shown in \Cref{tab:creative_results}, significantly reduces hint uses and total steps, with larger models benefiting the most. Key insights include:
\begin{itemize}[topsep=2pt, partopsep=-3pt, leftmargin=8pt, itemsep=-3pt]
\item Larger models still outperform smaller ones, suggesting while new modules aid creative reasoning, the core model's capabilities remain crucial.
\item Early exit progress improves across models, despite the exponentially increasing difficulty of consecutively making progress without hints, demonstrating EscapeAgent's effectiveness.
\end{itemize}


\begin{figure}[!t]
    \centering
    \subfigure{\includegraphics[width=\linewidth]{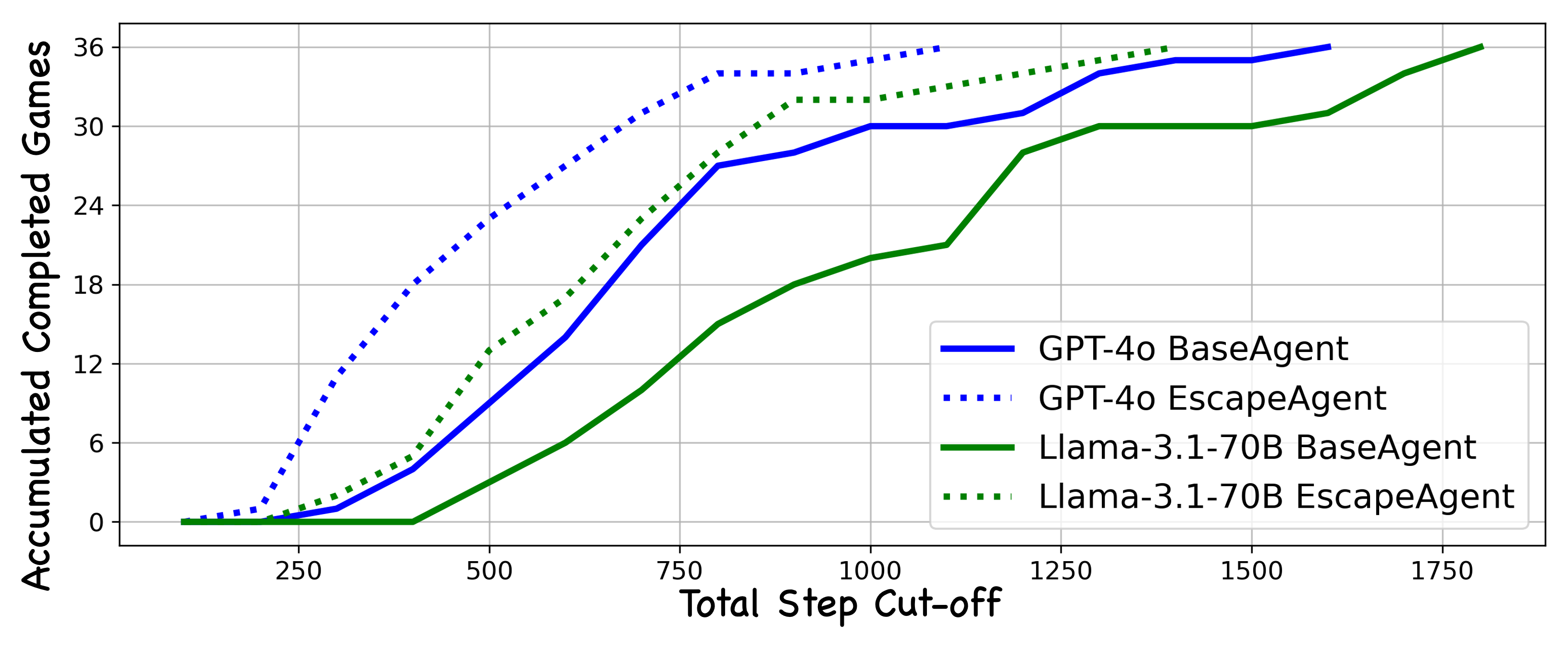}}
    \vspace{-5mm}
    \caption{The accumulated number of completed games (in total 36) relative to total steps a game setting takes. EscapeAgent, shown in dotted lines, completes games in fewer steps, demonstrating greater efficiency.}
    \label{fig:analysis_accumulate}
\end{figure}

\begin{figure*}[!t]
    \centering
    \subfigure{\includegraphics[width=\linewidth]{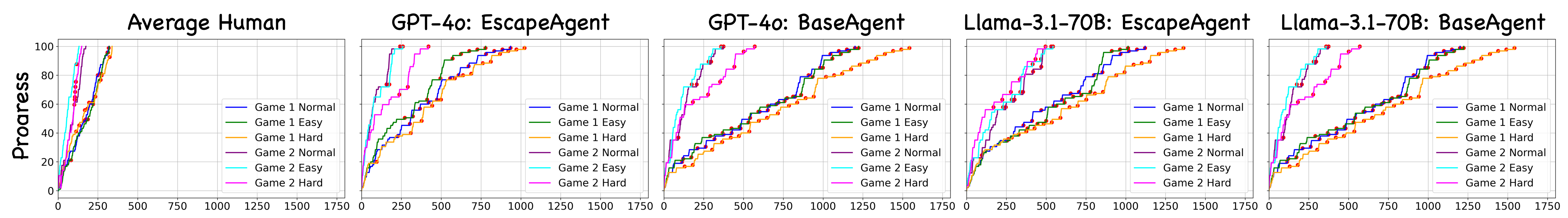}}
    \caption{Case study on Human, BaseAgent, and EscapeAgent's progress map corresponding to six game settings. Although EscapeAgent uses 40\% fewer hints and makes significant progress independently, it still falls far short of average human performance, often requiring twice as many total steps to complete a game.}
    \label{fig:analysis_progress}
\end{figure*}

\begin{figure}[!t]
    \centering
    \subfigure{\includegraphics[width=1\linewidth]{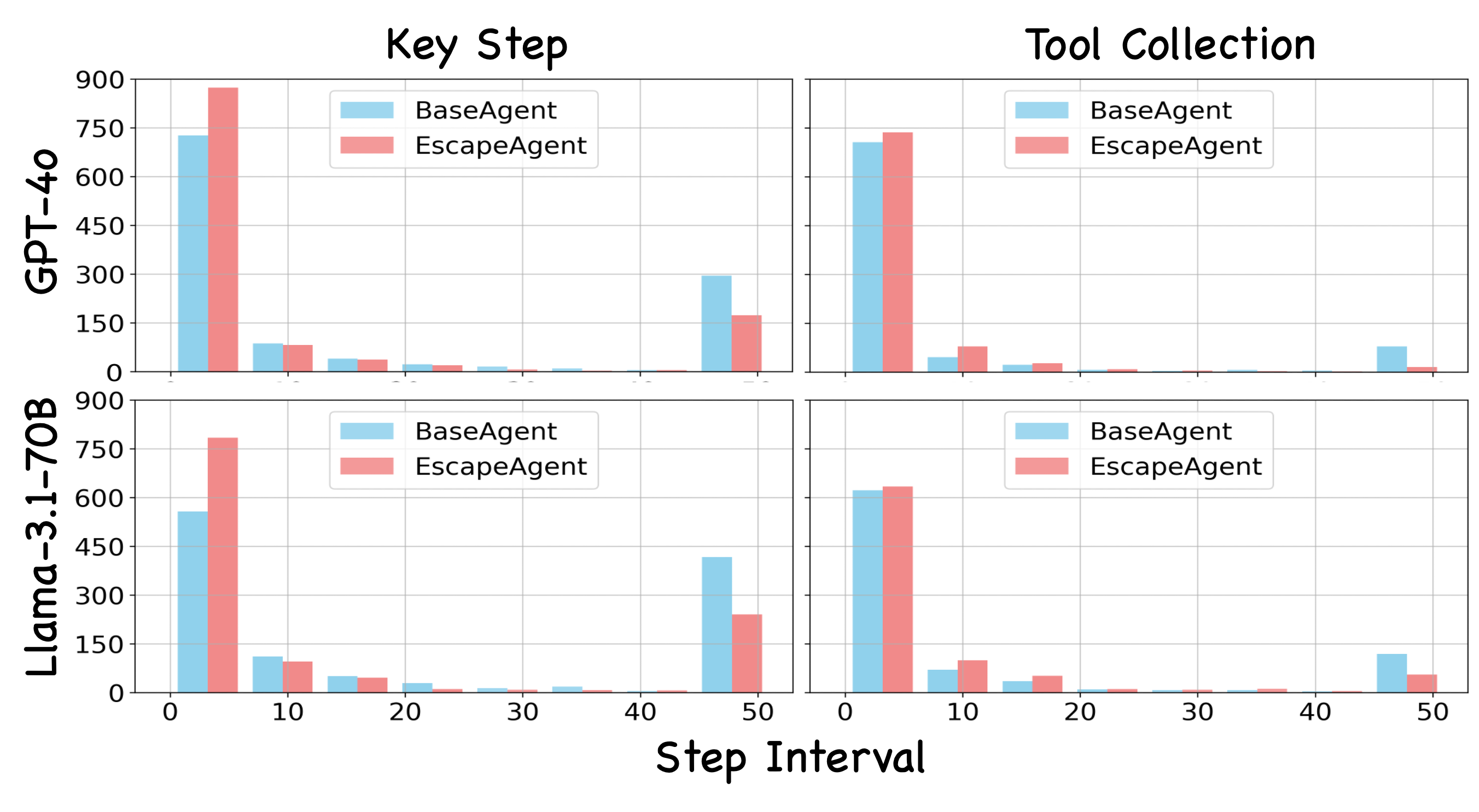}}
    \vspace{-8mm}
    \caption{Distribution of \textit{step intervals} for progress made through tool collection and key step achievement. EscapeAgent uses fewer steps to achieve the next progress and relies less on hints.}
    \label{fig:analysis_interval}
\end{figure}

\paragraph{EscapeAgent progresses more efficiently.}
As shown in \Cref{fig:analysis_accumulate}, EscapeAgent demonstrates steeper progress slopes, reflecting greater efficiency. \Cref{fig:analysis_interval} further shows that EscapeAgent requires fewer actions to reach the next key step, indicating stronger creative reasoning ability. The spike at 50 steps corresponds to hints provided after prolonged inactivity, so the lower pink bar here further highlights EscapeAgent's reduced hint dependency. Notably, across all models, progress after 15 steps without hints is rare, underscoring a lack of spontaneous insights typically seen in humans.

\paragraph{Case study.}
\Cref{fig:analysis_progress} illustrates agent progress relative to action steps across six game settings, with red dots marking hint provided. We observe that: \textbf{i)} EscapeAgent requires fewer hints and achieves steeper progress; \textbf{ii)} it can make consecutive progress in shorter intervals; \textbf{iii)} harder scenarios remain challenging, especially for BaseAgent, which heavily relies on hints; \textbf{iv)} Average human performs far better. Humans rarely make mistakes shown in \Cref{tab:error_analysis}, while agents still struggle with short memory due to context length and creative tool use strategies. These emphasize the need for further improvements and highlight EscapeBench's challenge to even the most advanced models.

\section{Further Analysis}

\subsection{Evaluation across Diverse Models}
\label{apdx:addition_diversity}

To enhance the diversity of model scales and domains in our evaluation, we present additional experimental results on smaller-scale and domain-specific models. These results supplement our main findings and provide further insights into the scalability and generalizability of our framework.

In addition to the models presented in out main table, we also evaluated sub-7B models, including the latest Llama-3.2-1B and 3B variants. However, these models demonstrated extremely limited performance. In even the simplest game scenarios under normal difficulty, these models frequently resorted to random guessing, with over 95\% of key actions requiring hint usage. The inefficiency translated into significantly inflated action sequences and minimal task completion, leading us to our conclusion that 7B is the minimum viable scale for meaningful evaluation in our benchmark.

\begin{table}[!t]
\centering
\resizebox{\linewidth}{!}{
\begin{tabular}{lccc}
\toprule
\textbf{Model} & \textbf{$^\downarrow$Hints Used} & \textbf{$^\downarrow$Total Steps} & \textbf{\makecell{$^{\uparrow}$Early Exit \\ Progress (\%)}} \\
\midrule
Qwen-2.5-7B-Instruct & 22.6 & \textbf{1329.8} & 5.93 \\
Qwen-2.5-Coder-7B    & \textbf{22.4} & 1339.8 & \textbf{12.99} \\
Qwen-2.5-Math-7B     & 45.4 & 2384.4 & 0.00 \\
\bottomrule
\end{tabular}
}
\caption{Performance of Domain-Specific Models on EscapeAgent Benchmark}
\label{tab:domain_models}
\end{table}

To examine domain specialization effects, we benchmarked two variants from the Qwen-2.5 model series, focused respectively on coding and mathematical tasks.
As shown in \Cref{tab:domain_models}, the Qwen-2.5-Coder-7B achieved performance close to the general-purpose 7B-Instruct model, indicating effective domain adaptation without significant loss of reasoning flexibility. In contrast, the Qwen-2.5-Math-7B model struggled, failing to initiate or complete key actions autonomously. This suggests that excessive specialization, such as strict math alignment, may hinder the model's ability to generalize and engage in creative, multi-step tasks.

\subsection{Ablation Study on Key Modules}
\label{sec:ablation_modules}

To evaluate the individual contributions of the \textit{Foresight} and \textit{Reflection} modules in EscapeAgent, we conducted an ablation study using two modified agent variants:
\begin{itemize}[topsep=2pt, partopsep=-3pt, leftmargin=8pt, itemsep=-3pt]
    \item \textbf{EscapeAgent (only Foresight):} Disables the task list maintained by the Reflection module, allowing action proposals based solely on environmental observation.
    \item \textbf{EscapeAgent (only Reflection):} Disables the Foresight module, relying instead on the task list to guide decision-making.
\end{itemize}

\begin{table}[!t]
\centering
\resizebox{\linewidth}{!}{
\begin{tabular}{lccc}
\toprule
\textbf{Model} & \textbf{$^\downarrow$Hints Used} & \textbf{$^\downarrow$Total Steps} & \textbf{\makecell{$^{\uparrow}$Early Exit \\ Progress (\%)}} \\
\midrule
GPT-4o, EscapeAgent & \textbf{5.00} & \textbf{452.00} & 44.55 \\
GPT-4o, only Reflection & 6.75 & 570.08 & 37.00 \\
GPT-4o, only Foresight & 7.17 & 593.92 & \textbf{48.89} \\
GPT-4o, BaseAgent & 9.83 & 707.33 & 19.85 \\
\midrule
Llama-3.1-70B, EscapeAgent & \textbf{7.17} & \textbf{624.67} & \textbf{28.31} \\
Llama-3.1-70B, only Reflection & 11.33 & 878.67 & 28.78 \\
Llama-3.1-70B, only Foresight & 11.00 & 877.25 & 22.29 \\
Llama-3.1-70B, BaseAgent & 14.67 & 981.42 & 17.28 \\
\bottomrule
\end{tabular}
}
\caption{Ablation results on GPT-4o and Llama-3.1-70B-Instruct backbones.}
\label{tab:ablation}
\end{table}

We benchmarked these variants alongside the full EscapeAgent and the BaseAgent across all game scenarios with normal difficulty. Across both GPT-4o and Llama-3.1-70B in \Cref{tab:ablation}, each module independently improves performance relative to the BaseAgent. The full EscapeAgent consistently achieves the best results, confirming that the Foresight and Reflection modules are complementary. Notably, Reflection alone often outperforms Foresight alone, likely because the Foresight module relies on the task list to avoid redundant or failed actions. These results support the conclusion that combining both modules is essential to fully realize EscapeAgent's potential.
\section{Discussions and Future Directions}
\label{discusstions}
\paragraph{LM's creativity for benchmarking.}
Our experiments reveal that even the most advanced language models within EscapeAgent require more hints and twice as many steps as the average human, exposing limitations in creative reasoning and tool use. The benchmark highlights that while analytical and practical intelligence is well-assessed, creative intelligence remains a critical gap. Addressing this gap may require enhancing LMs to link knowledge and objects through affordances—their properties and functions—to foster creativity.

\paragraph{Theoretical Foundations for AI Creativity.} Human creativity, characterized by generating novel ideas and adapting to complexity, arises from the interplay of stochastic neuronal noise and structured, learned information \cite{dainys2024human, malach2024neuronal}. In contrast, AI creativity relies on trained data patterns and algorithms. Boden identifies three mechanisms driving AI creativity: combining familiar ideas, exploring conceptual spaces, and enabling transformative innovations \cite{boden1998creativity}. Integrating insights from psychology and neuroscience may further enhance AI's creativity.

\paragraph{Human-AI Collaboration.} Human-AI collaboration in EscapeBench may promote a new problem-solving paradigm by merging human intuition with AI’s systematic reasoning. Humans bring unique insights and ideas that AI might not generate, while AI excels in tasks like information aggregation and logical organization. This synergy fosters innovative strategies, improves efficiency, and creates opportunities for deeper learning, offering a dynamic and enriched problem-solving experience that bridges human creativity with AI’s structured problem-solving capabilities.

\section{Conclusion}
In this work, we introduce EscapeBench, the first benchmark for advancing LM's creativity. Our results show that while LMs still lag in creative reasoning, the EscapeAgent framework improves innovative problem-solving and implicit goal identification. Despite these advancements, enhancing the models’ intrinsic creativity remains a challenge. Future work could explore integrating multi-modal perception and new reinforcement learning algorithms to foster greater creativity. Our work serves as an important first step, offering a robust environment for experimentation. Looking ahead, we believe that creative intelligence, beyond just analytical and practical capabilities, will play a key role in shaping the frontier of AI.

\section*{Limitations}
Our work utilizes a text-based environment to evaluate common language models, focusing on creative reasoning within this framework. However, an Escape Room scenario inherently includes visual and auditory clues, which we have not incorporated into this benchmark. Expanding to include multi-modal inputs could be a valuable next step for future work. Additionally, while the data used in our benchmark is annotated through intensive human effort to ensure high quality, this approach limits scalability. We have explored the use of GPT-4 for automatic annotation through free exploration but found that the model sometimes overlooks important items and clues, and struggles to design environment feedback crucial for adjusting the game’s difficulty. We anticipate that more powerful vision-language models may enable better automatic annotation in the future, though current model capabilities are still a limiting factor.

\section*{Ethical Statement}
In this research, we consider the following ethical issues related to our benchmark and agent design:\\
$\bullet$ \textbf{Fairness:} We ensure that EscapeBench is designed to provide equal evaluation opportunities for all agents, regardless of their underlying model architectures or training methodologies. The tasks and scenarios are crafted to assess creativity and problem-solving abilities without bias, promoting fairness in the benchmarking process. Additionally, we aim to avoid overfitting to specific agent strategies, ensuring a more generalizable and inclusive evaluation framework for future AI advancements. While our environment is robust, we caution against potential misuse and strongly encourage its fair and responsible use.\\
$\bullet$ \textbf{Transparency:} Our work incorporates Chain-of-Thought reasoning in the BaseAgent framework to improve the transparency and interpretability of the agent's decision-making process. This approach makes it easier to attribute the reasoning behind each agent's action. Additionally, we will fully release the benchmarking code, EscapeAgent design, and data to promote transparency in our evaluation process, ensuring that the broader research community can benefit from and build upon our work.

\section*{Acknowledgment}
This research is based upon work supported by U.S. DARPA ITM Program No. FA8650-23-C-7316 and DARPA ECOLE Program No. \#HR00112390060. The views and conclusions contained herein are those of the authors and should not be interpreted as necessarily representing the official policies, either expressed or implied, of DARPA, or the U.S. Government. The U.S. Government is authorized to reproduce and distribute reprints for governmental purposes notwithstanding any copyright annotation therein.

\bibliography{custom, ref-zdh}

\begin{thebibliography}{104}
\providecommand{\natexlab}[1]{#1}

\bibitem[{Abdin et~al.(2024)Abdin, Aneja, Awadalla, Awadallah, Awan, Bach, Bahree, Bakhtiari, Bao, Behl et~al.}]{abdin2024phi}
Marah Abdin, Jyoti Aneja, Hany Awadalla, Ahmed Awadallah, Ammar~Ahmad Awan, Nguyen Bach, Amit Bahree, Arash Bakhtiari, Jianmin Bao, Harkirat Behl, et~al. 2024.
\newblock Phi-3 technical report: A highly capable language model locally on your phone.
\newblock \emph{arXiv preprint arXiv:2404.14219}.

\bibitem[{Agashe et~al.(2023)Agashe, Fan, and Wang}]{multi_agent_coordination}
Saaket Agashe, Yue Fan, and Xin~Eric Wang. 2023.
\newblock Evaluating multi-agent coordination abilities in large language models.
\newblock \emph{arXiv preprint arXiv:2310.03903}.

\bibitem[{Akoury et~al.(2020)Akoury, Wang, Whiting, Hood, Peng, and Iyyer}]{akoury2020storium}
Nader Akoury, Shufan Wang, Josh Whiting, Stephen Hood, Nanyun Peng, and Mohit Iyyer. 2020.
\newblock Storium: A dataset and evaluation platform for machine-in-the-loop story generation.
\newblock \emph{arXiv preprint arXiv:2010.01717}.

\bibitem[{{Anthropic}(2024)}]{2024claude}
{Anthropic}. 2024.
\newblock Introducing claude 3.5 sonnet.
\newblock \url{https://www.anthropic.com/news/claude-3-5-sonnet}.

\bibitem[{BAAI(2023)}]{baai2023plan4mc}
PKU BAAI. 2023.
\newblock Plan4mc: Skill reinforcement learning and planning for open-world minecraft tasks.
\newblock \emph{arXiv preprint arXiv:2303.16563}.

\bibitem[{Boden(1998)}]{boden1998creativity}
Margaret~A. Boden. 1998.
\newblock Creativity and artificial intelligence.
\newblock \emph{Artificial Intelligence}, 103(1-2):347--356.

\bibitem[{Boiko et~al.(2023)Boiko, MacKnight, and Gomes}]{boiko2023emergent}
Daniil~A Boiko, Robert MacKnight, and Gabe Gomes. 2023.
\newblock Emergent autonomous scientific research capabilities of large language models.
\newblock \emph{arXiv preprint arXiv:2304.05332}.

\bibitem[{Bran et~al.(2023)Bran, Cox, Schilter, Baldassari, White, and Schwaller}]{bran2023chemcrow}
Andres~M Bran, Sam Cox, Oliver Schilter, Carlo Baldassari, Andrew~D White, and Philippe Schwaller. 2023.
\newblock Chemcrow: Augmenting large-language models with chemistry tools.
\newblock \emph{arXiv preprint arXiv:2304.05376}.

\bibitem[{Brown et~al.(2020)}]{brown2020language}
Tom Brown et~al. 2020.
\newblock Language models are few-shot learners.
\newblock \emph{Advances in neural information processing systems}.

\bibitem[{Cai et~al.(2024)Cai, Wang, Ma, Chen, and Zhou}]{cai2023large}
Tianle Cai, Xuezhi Wang, Tengyu Ma, Xinyun Chen, and Denny Zhou. 2024.
\newblock \href {https://openreview.net/forum?id=qV83K9d5WB} {Large language models as tool makers}.
\newblock In \emph{The Twelfth International Conference on Learning Representations}.

\bibitem[{Carroll et~al.(2019)Carroll, Shah, Ho, Griffiths, Seshia, Abbeel, and Dragan}]{carroll2019utility}
Micah Carroll, Rohin Shah, Mark~K Ho, Tom Griffiths, Sanjit Seshia, Pieter Abbeel, and Anca Dragan. 2019.
\newblock On the utility of learning about humans for human-ai coordination.
\newblock \emph{Advances in neural information processing systems}, 32.

\bibitem[{Chen et~al.(2022)Chen, Jiang, Jin, Zhou, Liu, Brantingham, and Wang}]{chen2022reliable}
Xiusi Chen, Jyun-Yu Jiang, Kun Jin, Yichao Zhou, Mingyan Liu, P~Jeffrey Brantingham, and Wei Wang. 2022.
\newblock Reliable: Offline reinforcement learning for tactical strategies in professional basketball games.
\newblock In \emph{Proceedings of the 31st ACM International Conference on Information \& Knowledge Management}, pages 3023--3032.

\bibitem[{Chen et~al.(2024)Chen, Wang, Hu, Reynoso, Jin, Liu, Brantingham, and Wang}]{chen2024playbest}
Xiusi Chen, Wei-Yao Wang, Ziniu Hu, David Reynoso, Kun Jin, Mingyan Liu, P~Jeffrey Brantingham, and Wei Wang. 2024.
\newblock Playbest: Professional basketball player behavior synthesis via planning with diffusion.
\newblock In \emph{Proceedings of the 33rd ACM International Conference on Information and Knowledge Management}, pages 4406--4413.

\bibitem[{C{\^o}t{\'e} et~al.(2019)C{\^o}t{\'e}, K{\'a}d{\'a}r, Yuan, Kybartas, Barnes, Fine, Moore, Hausknecht, El~Asri, Adada et~al.}]{cote2019textworld}
Marc-Alexandre C{\^o}t{\'e}, Akos K{\'a}d{\'a}r, Xingdi Yuan, Ben Kybartas, Tavian Barnes, Emery Fine, James Moore, Matthew Hausknecht, Layla El~Asri, Mahmoud Adada, et~al. 2019.
\newblock Textworld: A learning environment for text-based games.
\newblock In \emph{Computer Games: 7th Workshop, CGW 2018, Held in Conjunction with the 27th International Conference on Artificial Intelligence, IJCAI 2018, Stockholm, Sweden, July 13, 2018, Revised Selected Papers 7}, pages 41--75. Springer.

\bibitem[{Dainys(2024)}]{dainys2024human}
Augustinas Dainys. 2024.
\newblock Human creativity versus machine creativity: Will humans be surpassed by ai?

\bibitem[{Deng et~al.(2024)Deng, Gu, Zheng, Chen, Stevens, Wang, Sun, and Su}]{deng2024mind2web}
Xiang Deng, Yu~Gu, Boyuan Zheng, Shijie Chen, Sam Stevens, Boshi Wang, Huan Sun, and Yu~Su. 2024.
\newblock Mind2web: Towards a generalist agent for the web.
\newblock \emph{Advances in Neural Information Processing Systems}, 36.

\bibitem[{Dhuliawala et~al.(2024)Dhuliawala, Komeili, Xu, Raileanu, Li, Celikyilmaz, and Weston}]{dhuliawala2024chain}
Shehzaad Dhuliawala, Mojtaba Komeili, Jing Xu, Roberta Raileanu, Xian Li, Asli Celikyilmaz, and Jason~E Weston. 2024.
\newblock Chain-of-verification reduces hallucination in large language models.
\newblock In \emph{ICLR 2024 Workshop on Reliable and Responsible Foundation Models}.

\bibitem[{Dubey et~al.(2024)Dubey, Jauhri, Pandey, Kadian, Al-Dahle, Letman, Mathur, Schelten, Yang, Fan et~al.}]{dubey2024llama}
Abhimanyu Dubey, Abhinav Jauhri, Abhinav Pandey, Abhishek Kadian, Ahmad Al-Dahle, Aiesha Letman, Akhil Mathur, Alan Schelten, Amy Yang, Angela Fan, et~al. 2024.
\newblock The llama 3 herd of models.
\newblock \emph{arXiv preprint arXiv:2407.21783}.

\bibitem[{Fan et~al.(2022)Fan, Wang, Jiang, Mandlekar, Yang, Zhu, Tang, Huang, Zhu, and Anandkumar}]{fan2022minedojo}
Linxi Fan, Guanzhi Wang, Yunfan Jiang, Ajay Mandlekar, Yuncong Yang, Haoyi Zhu, Andrew Tang, De-An Huang, Yuke Zhu, and Anima Anandkumar. 2022.
\newblock Minedojo: Building open-ended embodied agents with internet-scale knowledge.
\newblock \emph{Advances in Neural Information Processing Systems}, 35:18343--18362.

\bibitem[{Franceschelli and Musolesi(2023)}]{franceschelli2023creativity}
Giorgio Franceschelli and Mirco Musolesi. 2023.
\newblock On the creativity of large language models.
\newblock \emph{arXiv preprint arXiv:2304.00008}.

\bibitem[{Furuta et~al.(2024)Furuta, Lee, Nachum, Matsuo, Faust, Gu, and Gur}]{furuta2024multimodal}
Hiroki Furuta, Kuang-Huei Lee, Ofir Nachum, Yutaka Matsuo, Aleksandra Faust, Shixiang~Shane Gu, and Izzeddin Gur. 2024.
\newblock Multimodal web navigation with instruction-finetuned foundation models.
\newblock In \emph{The Twelfth International Conference on Learning Representations}.

\bibitem[{Gan et~al.(2021)Gan, Zhou, Schwartz, Alter, Bhandwaldar, Gutfreund, Yamins, DiCarlo, McDermott, Torralba et~al.}]{gan2021threedworld}
Chuang Gan, Siyuan Zhou, Jeremy Schwartz, Seth Alter, Abhishek Bhandwaldar, Dan Gutfreund, Daniel~LK Yamins, James~J DiCarlo, Josh McDermott, Antonio Torralba, et~al. 2021.
\newblock The threedworld transport challenge: A visually guided task-and-motion planning benchmark for physically realistic embodied ai.
\newblock \emph{arXiv preprint arXiv:2103.14025}.

\bibitem[{Guilford(1967)}]{guilford1967creativity}
Joy~P Guilford. 1967.
\newblock Creativity: Yesterday, today and tomorrow.
\newblock \emph{The Journal of Creative Behavior}, 1(1):3--14.

\bibitem[{Guo et~al.(2023)Guo, Yang, Yoo, Lin, Iwasawa, and Matsuo}]{guo2023suspicion}
Jiaxian Guo, Bo~Yang, Paul Yoo, Bill~Yuchen Lin, Yusuke Iwasawa, and Yutaka Matsuo. 2023.
\newblock Suspicion-agent: Playing imperfect information games with theory of mind aware gpt-4.
\newblock \emph{arXiv preprint arXiv:2309.17277}.

\bibitem[{Gur et~al.(2024)Gur, Furuta, Huang, Safdari, Matsuo, Eck, and Faust}]{gur2024real}
Izzeddin Gur, Hiroki Furuta, Austin~V Huang, Mustafa Safdari, Yutaka Matsuo, Douglas Eck, and Aleksandra Faust. 2024.
\newblock A real-world webagent with planning, long context understanding, and program synthesis.
\newblock In \emph{The Twelfth International Conference on Learning Representations}.

\bibitem[{Guzik et~al.(2023)Guzik, Byrge, and Gilde}]{GUZIK2023100065}
Erik~E. Guzik, Christian Byrge, and Christian Gilde. 2023.
\newblock \href {https://doi.org/10.1016/j.yjoc.2023.100065} {The originality of machines: Ai takes the torrance test}.
\newblock \emph{Journal of Creativity}, 33(3):100065.

\bibitem[{Hao et~al.(2023)Hao, Gu, Ma, Hong, Wang, Wang, and Hu}]{hao2023reasoning}
Shibo Hao, Yi~Gu, Haodi Ma, Joshua Hong, Zhen Wang, Daisy Wang, and Zhiting Hu. 2023.
\newblock Reasoning with language model is planning with world model.
\newblock In \emph{Proceedings of the 2023 Conference on Empirical Methods in Natural Language Processing}, pages 8154--8173.

\bibitem[{Hu et~al.(2023)Hu, Fu, Du, Luo, Zhao, and Zhao}]{hu2023chatdb}
Chenxu Hu, Jie Fu, Chenzhuang Du, Simian Luo, Junbo Zhao, and Hang Zhao. 2023.
\newblock Chatdb: Augmenting llms with databases as their symbolic memory.
\newblock \emph{arXiv preprint arXiv:2306.03901}.

\bibitem[{Hu et~al.(2024)Hu, Huang, Ilhan, Tekin, Liu, Kompella, and Liu}]{hu2024survey}
Sihao Hu, Tiansheng Huang, Fatih Ilhan, Selim Tekin, Gaowen Liu, Ramana Kompella, and Ling Liu. 2024.
\newblock A survey on large language model-based game agents.
\newblock \emph{arXiv preprint arXiv:2404.02039}.

\bibitem[{Huang et~al.(2024{\natexlab{a}})Huang, Yong, Ma, Linghu, Li, Wang, Li, Zhu, Jia, and Huang}]{huang2024embodied}
Jiangyong Huang, Silong Yong, Xiaojian Ma, Xiongkun Linghu, Puhao Li, Yan Wang, Qing Li, Song-Chun Zhu, Baoxiong Jia, and Siyuan Huang. 2024{\natexlab{a}}.
\newblock An embodied generalist agent in 3d world.
\newblock In \emph{Forty-first International Conference on Machine Learning}.

\bibitem[{Huang et~al.(2023{\natexlab{a}})Huang, Vora, Liang, and Leskovec}]{huang2023benchmarking}
Qian Huang, Jian Vora, Percy Liang, and Jure Leskovec. 2023{\natexlab{a}}.
\newblock Benchmarking large language models as ai research agents.
\newblock In \emph{NeurIPS 2023 Foundation Models for Decision Making Workshop}.

\bibitem[{Huang et~al.(2024{\natexlab{b}})Huang, Wang, Li, Zhang, and Fei-Fei}]{huang2024rekep}
Wenlong Huang, Chen Wang, Yunzhu Li, Ruohan Zhang, and Li~Fei-Fei. 2024{\natexlab{b}}.
\newblock Rekep: Spatio-temporal reasoning of relational keypoint constraints for robotic manipulation.
\newblock \emph{arXiv preprint arXiv:2409.01652}.

\bibitem[{Huang et~al.(2023{\natexlab{b}})Huang, Wang, Zhang, Li, Wu, and Fei-Fei}]{huang2023voxposer}
Wenlong Huang, Chen Wang, Ruohan Zhang, Yunzhu Li, Jiajun Wu, and Li~Fei-Fei. 2023{\natexlab{b}}.
\newblock Voxposer: Composable 3d value maps for robotic manipulation with language models.
\newblock \emph{arXiv preprint arXiv:2307.05973}.

\bibitem[{Hurst et~al.(2024)Hurst, Lerer, Goucher, Perelman, Ramesh, Clark, Ostrow, Welihinda, Hayes, Radford et~al.}]{hurst2024gpt}
Aaron Hurst, Adam Lerer, Adam~P Goucher, Adam Perelman, Aditya Ramesh, Aidan Clark, AJ~Ostrow, Akila Welihinda, Alan Hayes, Alec Radford, et~al. 2024.
\newblock Gpt-4o system card.
\newblock \emph{arXiv preprint arXiv:2410.21276}.

\bibitem[{{Infocom}(1980)}]{zork1980}
{Infocom}. 1980.
\newblock {Zork I}.
\newblock \url{http://ifdb.tads.org/viewgame?id=0dbnusxunq7fw5ro}.

\bibitem[{Kojima et~al.(2022)Kojima, Gu, Reid, Matsuo, and Iwasawa}]{kojima2022large}
Takeshi Kojima, Shixiang~Shane Gu, Machel Reid, Yutaka Matsuo, and Yusuke Iwasawa. 2022.
\newblock Large language models are zero-shot reasoners.
\newblock \emph{Advances in neural information processing systems}, 35:22199--22213.

\bibitem[{Lake et~al.(2017)Lake, Ullman, Tenenbaum, and Gershman}]{lake2017building}
Brenden~M Lake, Tomer~D Ullman, Joshua~B Tenenbaum, and Samuel~J Gershman. 2017.
\newblock Building machines that learn and think like people.
\newblock \emph{Behavioral and brain sciences}, 40:e253.

\bibitem[{Legg and Hutter(2007)}]{legg2007universal}
Shane Legg and Marcus Hutter. 2007.
\newblock Universal intelligence: A definition of machine intelligence.
\newblock \emph{Minds and machines}, 17:391--444.

\bibitem[{Li et~al.(2023)Li, Hammoud, Itani, Khizbullin, and Ghanem}]{li2023camel}
Guohao Li, Hasan Hammoud, Hani Itani, Dmitrii Khizbullin, and Bernard Ghanem. 2023.
\newblock Camel: Communicative agents for" mind" exploration of large language model society.
\newblock \emph{Advances in Neural Information Processing Systems}, 36:51991--52008.

\bibitem[{Liang et~al.(2023{\natexlab{a}})Liang, Huang, Xia, Xu, Hausman, Ichter, Florence, and Zeng}]{liang2023code}
Jacky Liang, Wenlong Huang, Fei Xia, Peng Xu, Karol Hausman, Brian Ichter, Pete Florence, and Andy Zeng. 2023{\natexlab{a}}.
\newblock Code as policies: Language model programs for embodied control.
\newblock In \emph{2023 IEEE International Conference on Robotics and Automation (ICRA)}, pages 9493--9500. IEEE.

\bibitem[{Liang et~al.(2023{\natexlab{b}})Liang, Wang, Huang, Wu, Wu, Lu, Ma, and Li}]{liang2023unleashing}
Xinnian Liang, Bing Wang, Hui Huang, Shuangzhi Wu, Peihao Wu, Lu~Lu, Zejun Ma, and Zhoujun Li. 2023{\natexlab{b}}.
\newblock Unleashing infinite-length input capacity for large-scale language models with self-controlled memory system.
\newblock \emph{arXiv e-prints}, pages arXiv--2304.

\bibitem[{Lin et~al.(2024)Lin, Fu, Yang, Brahman, Huang, Bhagavatula, Ammanabrolu, Choi, and Ren}]{lin2024swiftsage}
Bill~Yuchen Lin, Yicheng Fu, Karina Yang, Faeze Brahman, Shiyu Huang, Chandra Bhagavatula, Prithviraj Ammanabrolu, Yejin Choi, and Xiang Ren. 2024.
\newblock Swiftsage: A generative agent with fast and slow thinking for complex interactive tasks.
\newblock \emph{Advances in Neural Information Processing Systems}, 36.

\bibitem[{Lin et~al.(2023)Lin, Zhao, Zhang, Wu, Ping, and Chen}]{lin2023agentsims}
Jiaju Lin, Haoran Zhao, Aochi Zhang, Yiting Wu, Huqiuyue Ping, and Qin Chen. 2023.
\newblock Agentsims: An open-source sandbox for large language model evaluation.
\newblock \emph{arXiv preprint arXiv:2308.04026}.

\bibitem[{Liu et~al.(2024{\natexlab{a}})Liu, Feng, Wang, Wang, Liu, Zhao, Dengr, Ruan, Dai, Guo et~al.}]{liu2024deepseek}
Aixin Liu, Bei Feng, Bin Wang, Bingxuan Wang, Bo~Liu, Chenggang Zhao, Chengqi Dengr, Chong Ruan, Damai Dai, Daya Guo, et~al. 2024{\natexlab{a}}.
\newblock Deepseek-v2: A strong, economical, and efficient mixture-of-experts language model.
\newblock \emph{arXiv preprint arXiv:2405.04434}.

\bibitem[{Liu et~al.(2023{\natexlab{a}})Liu, Jiang, Zhang, Liu, Zhang, Biswas, and Stone}]{liu2023llm+}
Bo~Liu, Yuqian Jiang, Xiaohan Zhang, Qiang Liu, Shiqi Zhang, Joydeep Biswas, and Peter Stone. 2023{\natexlab{a}}.
\newblock Llm+ p: Empowering large language models with optimal planning proficiency.
\newblock \emph{arXiv preprint arXiv:2304.11477}.

\bibitem[{Liu et~al.(2023{\natexlab{b}})Liu, Yang, Shen, Hu, Zhang, Gu, and Zhang}]{liu2023think}
Lei Liu, Xiaoyan Yang, Yue Shen, Binbin Hu, Zhiqiang Zhang, Jinjie Gu, and Guannan Zhang. 2023{\natexlab{b}}.
\newblock Think-in-memory: Recalling and post-thinking enable llms with long-term memory.
\newblock \emph{arXiv preprint arXiv:2311.08719}.

\bibitem[{Liu et~al.(2023{\natexlab{c}})Liu, Yang, Jia, Zhang, Zhou, Dai, Yang, and Vosoughi}]{liu2023training}
Ruibo Liu, Ruixin Yang, Chenyan Jia, Ge~Zhang, Denny Zhou, Andrew~M Dai, Diyi Yang, and Soroush Vosoughi. 2023{\natexlab{c}}.
\newblock Training socially aligned language models in simulated human society.
\newblock \emph{arXiv preprint arXiv:2305.16960}.

\bibitem[{Liu et~al.(2024{\natexlab{b}})Liu, Yu, Zhang, Xu, Lei, Lai, Gu, Ding, Men, Yang et~al.}]{liu2024agentbench}
Xiao Liu, Hao Yu, Hanchen Zhang, Yifan Xu, Xuanyu Lei, Hanyu Lai, Yu~Gu, Hangliang Ding, Kaiwen Men, Kejuan Yang, et~al. 2024{\natexlab{b}}.
\newblock Agentbench: Evaluating llms as agents.
\newblock In \emph{The Twelfth International Conference on Learning Representations}.

\bibitem[{Madaan et~al.(2024)Madaan, Tandon, Gupta, Hallinan, Gao, Wiegreffe, Alon, Dziri, Prabhumoye, Yang et~al.}]{madaan2024self}
Aman Madaan, Niket Tandon, Prakhar Gupta, Skyler Hallinan, Luyu Gao, Sarah Wiegreffe, Uri Alon, Nouha Dziri, Shrimai Prabhumoye, Yiming Yang, et~al. 2024.
\newblock Self-refine: Iterative refinement with self-feedback.
\newblock \emph{Advances in Neural Information Processing Systems}, 36.

\bibitem[{Malach(2024)}]{malach2024neuronal}
Rafael Malach. 2024.
\newblock The neuronal basis of human creativity.
\newblock \emph{Frontiers in Human Neuroscience}, 18:1367922.

\bibitem[{Mialon et~al.(2023)Mialon, Dess{\`\i}, Lomeli, Nalmpantis, Pasunuru, Raileanu, Rozi{\`e}re, Schick, Dwivedi-Yu, Celikyilmaz et~al.}]{mialon2023augmented}
Gr{\'e}goire Mialon, Roberto Dess{\`\i}, Maria Lomeli, Christoforos Nalmpantis, Ram Pasunuru, Roberta Raileanu, Baptiste Rozi{\`e}re, Timo Schick, Jane Dwivedi-Yu, Asli Celikyilmaz, et~al. 2023.
\newblock Augmented language models: a survey.
\newblock \emph{arXiv preprint arXiv:2302.07842}.

\bibitem[{Miao et~al.(2024)Miao, Teh, and Rainforth}]{miao2024selfcheck}
Ning Miao, Yee~Whye Teh, and Tom Rainforth. 2024.
\newblock Selfcheck: Using llms to zero-shot check their own step-by-step reasoning.
\newblock In \emph{The Twelfth International Conference on Learning Representations}.

\bibitem[{{MistralAI}(2024)}]{2024ministral}
{MistralAI}. 2024.
\newblock Introducing ministral-8b-instruct.
\newblock \url{https://huggingface.co/mistralai/Ministral-8B-Instruct-2410}.

\bibitem[{Nakano et~al.(2021)Nakano, Hilton, Balaji, Wu, Ouyang, Kim, Hesse, Jain, Kosaraju, Saunders et~al.}]{nakano2021webgpt}
Reiichiro Nakano, Jacob Hilton, Suchir Balaji, Jeff Wu, Long Ouyang, Christina Kim, Christopher Hesse, Shantanu Jain, Vineet Kosaraju, William Saunders, et~al. 2021.
\newblock Webgpt: Browser-assisted question-answering with human feedback.
\newblock \emph{arXiv preprint arXiv:2112.09332}.

\bibitem[{O'Gara(2023)}]{o2023hoodwinked}
Aidan O'Gara. 2023.
\newblock Hoodwinked: Deception and cooperation in a text-based game for language models.
\newblock \emph{arXiv preprint arXiv:2308.01404}.

\bibitem[{Park et~al.(2023)Park, O'Brien, Cai, Morris, Liang, and Bernstein}]{park2023generative}
Joon~Sung Park, Joseph O'Brien, Carrie~Jun Cai, Meredith~Ringel Morris, Percy Liang, and Michael~S Bernstein. 2023.
\newblock Generative agents: Interactive simulacra of human behavior.
\newblock In \emph{Proceedings of the 36th annual acm symposium on user interface software and technology}, pages 1--22.

\bibitem[{Qi et~al.(2024)Qi, Chen, Li, Kong, Wang, Yang, Wong, Zhong, Zhang, Zhang et~al.}]{qi2024civrealm}
Siyuan Qi, Shuo Chen, Yexin Li, Xiangyu Kong, Junqi Wang, Bangcheng Yang, Pring Wong, Yifan Zhong, Xiaoyuan Zhang, Zhaowei Zhang, et~al. 2024.
\newblock Civrealm: A learning and reasoning odyssey in civilization for decision-making agents.
\newblock \emph{arXiv preprint arXiv:2401.10568}.

\bibitem[{Qian et~al.(2025{\natexlab{a}})Qian, Acikgoz, He, Wang, Chen, Hakkani-T{\"u}r, Tur, and Ji}]{qian2025toolrl}
Cheng Qian, Emre~Can Acikgoz, Qi~He, Hongru Wang, Xiusi Chen, Dilek Hakkani-T{\"u}r, Gokhan Tur, and Heng Ji. 2025{\natexlab{a}}.
\newblock Toolrl: Reward is all tool learning needs.
\newblock \emph{arXiv preprint arXiv:2504.13958}.

\bibitem[{Qian et~al.(2025{\natexlab{b}})Qian, Du, Wang, Chen, Zhang, Sil, Zhai, McKeown, and Ji}]{qian2025modelingagent}
Cheng Qian, Hongyi Du, Hongru Wang, Xiusi Chen, Yuji Zhang, Avirup Sil, Chengxiang Zhai, Kathleen McKeown, and Heng Ji. 2025{\natexlab{b}}.
\newblock Modelingagent: Bridging llms and mathematical modeling for real-world challenges.
\newblock \emph{arXiv preprint arXiv:2505.15068}.

\bibitem[{Qian et~al.(2023)Qian, Han, Fung, Qin, Liu, and Ji}]{qian2023creator}
Cheng Qian, Chi Han, Yi~Fung, Yujia Qin, Zhiyuan Liu, and Heng Ji. 2023.
\newblock Creator: Tool creation for disentangling abstract and concrete reasoning of large language models.
\newblock In \emph{Findings of the Association for Computational Linguistics: EMNLP 2023}, pages 6922--6939.

\bibitem[{Qian et~al.(2024)Qian, Xiong, Liu, and Liu}]{qian2024toolink}
Cheng Qian, Chenyan Xiong, Zhenghao Liu, and Zhiyuan Liu. 2024.
\newblock Toolink: Linking toolkit creation and using through chain-of-solving on open-source model.
\newblock In \emph{Proceedings of the 2024 Conference of the North American Chapter of the Association for Computational Linguistics: Human Language Technologies (Volume 1: Long Papers)}, pages 831--854.

\bibitem[{Qiao et~al.(2023)Qiao, Wu, Liang, Li, and Duan}]{GameEval}
Dan Qiao, Chenfei Wu, Yaobo Liang, Juntao Li, and Nan Duan. 2023.
\newblock Gameeval: Evaluating llms on conversational games.
\newblock \emph{arXiv preprint arXiv:2308.10032}.

\bibitem[{Qin et~al.(2023)Qin, Hu, Lin, Chen, Ding, Cui, Zeng, Huang, Xiao, Han et~al.}]{qin2023tool}
Yujia Qin, Shengding Hu, Yankai Lin, Weize Chen, Ning Ding, Ganqu Cui, Zheni Zeng, Yufei Huang, Chaojun Xiao, Chi Han, et~al. 2023.
\newblock Tool learning with foundation models.
\newblock \emph{arXiv preprint arXiv.2304.08354}, 10.

\bibitem[{Rana et~al.(2023)Rana, Haviland, Garg, Abou-Chakra, Reid, and Suenderhauf}]{rana2023sayplan}
Krishan Rana, Jesse Haviland, Sourav Garg, Jad Abou-Chakra, Ian~D Reid, and Niko Suenderhauf. 2023.
\newblock Sayplan: Grounding large language models using 3d scene graphs for scalable task planning.
\newblock \emph{CoRR}.

\bibitem[{Ren et~al.(2024)Ren, Cui, Song, Wang, and Hu}]{ren2024emergence}
Siyue Ren, Zhiyao Cui, Ruiqi Song, Zhen Wang, and Shuyue Hu. 2024.
\newblock Emergence of social norms in large language model-based agent societies.
\newblock \emph{arXiv preprint arXiv:2403.08251}.

\bibitem[{Rospigliosi(2022)}]{rospigliosi2022metaverse}
Pericles~‘asher’ Rospigliosi. 2022.
\newblock Metaverse or simulacra? roblox, minecraft, meta and the turn to virtual reality for education, socialisation and work.

\bibitem[{Schick et~al.(2023)Schick, Dwivedi-Yu, Dessi, Raileanu, Lomeli, Hambro, Zettlemoyer, Cancedda, and Scialom}]{schick2023toolformer}
Timo Schick, Jane Dwivedi-Yu, Roberto Dessi, Roberta Raileanu, Maria Lomeli, Eric Hambro, Luke Zettlemoyer, Nicola Cancedda, and Thomas Scialom. 2023.
\newblock Toolformer: Language models can teach themselves to use tools.
\newblock In \emph{Thirty-seventh Conference on Neural Information Processing Systems}.

\bibitem[{Si et~al.(2024)Si, Yang, and Hashimoto}]{si2024can}
Chenglei Si, Diyi Yang, and Tatsunori Hashimoto. 2024.
\newblock Can llms generate novel research ideas? a large-scale human study with 100+ nlp researchers.
\newblock \emph{arXiv preprint arXiv:2409.04109}.

\bibitem[{Sternberg(1984)}]{sternberg1984toward}
Robert~J Sternberg. 1984.
\newblock Toward a triarchic theory of human intelligence.
\newblock \emph{Behavioral and Brain Sciences}, 7(2):269--287.

\bibitem[{Team et~al.(2024)Team, Georgiev, Lei, Burnell, Bai, Gulati, Tanzer, Vincent, Pan, Wang et~al.}]{team2024gemini}
Gemini Team, Petko Georgiev, Ving~Ian Lei, Ryan Burnell, Libin Bai, Anmol Gulati, Garrett Tanzer, Damien Vincent, Zhufeng Pan, Shibo Wang, et~al. 2024.
\newblock Gemini 1.5: Unlocking multimodal understanding across millions of tokens of context.
\newblock \emph{arXiv preprint arXiv:2403.05530}.

\bibitem[{Team(2024)}]{2024qwen2.5}
Qwen Team. 2024.
\newblock \href {https://qwenlm.github.io/blog/qwen2.5/} {Qwen2.5: A party of foundation models}.

\bibitem[{Uluda{\u{g}}l{\i} and O{\u{g}}uz(2023)}]{uludaugli2023non}
Muhtar~{\c{C}}a{\u{g}}kan Uluda{\u{g}}l{\i} and Kaya O{\u{g}}uz. 2023.
\newblock Non-player character decision-making in computer games.
\newblock \emph{Artificial Intelligence Review}, 56(12):14159--14191.

\bibitem[{Urbanek et~al.(2019)Urbanek, Fan, Karamcheti, Jain, Humeau, Dinan, Rockt{\"a}schel, Kiela, Szlam, and Weston}]{urbanek2019learning}
Jack Urbanek, Angela Fan, Siddharth Karamcheti, Saachi Jain, Samuel Humeau, Emily Dinan, Tim Rockt{\"a}schel, Douwe Kiela, Arthur Szlam, and Jason Weston. 2019.
\newblock Learning to speak and act in a fantasy text adventure game.
\newblock In \emph{Proceedings of the 2019 Conference on Empirical Methods in Natural Language Processing and the 9th International Joint Conference on Natural Language Processing (EMNLP-IJCNLP)}, pages 673--683.

\bibitem[{Wang et~al.(2024{\natexlab{a}})Wang, Xie, Jiang, Mandlekar, Xiao, Zhu, Fan, and Anandkumar}]{wang2024voyager}
Guanzhi Wang, Yuqi Xie, Yunfan Jiang, Ajay Mandlekar, Chaowei Xiao, Yuke Zhu, Linxi Fan, and Anima Anandkumar. 2024{\natexlab{a}}.
\newblock Voyager: An open-ended embodied agent with large language models.
\newblock \emph{Transactions on Machine Learning Research}.

\bibitem[{Wang et~al.(2025)Wang, Qian, Zhong, Chen, Qiu, Huang, Jin, Wang, Wong, and Ji}]{wang2025otc}
Hongru Wang, Cheng Qian, Wanjun Zhong, Xiusi Chen, Jiahao Qiu, Shijue Huang, Bowen Jin, Mengdi Wang, Kam-Fai Wong, and Heng Ji. 2025.
\newblock Otc: Optimal tool calls via reinforcement learning.
\newblock \emph{arXiv preprint arXiv:2504.14870}.

\bibitem[{Wang et~al.(2024{\natexlab{b}})Wang, Downey, Ji, and Hope}]{scimon2024}
Qingyun Wang, Doug Downey, Heng Ji, and Tom Hope. 2024{\natexlab{b}}.
\newblock Scimon: Scientific inspiration machines optimized for novelty.
\newblock In \emph{Proc. The 62nd Annual Meeting of the Association for Computational Linguistics (ACL2024)}.

\bibitem[{Wang et~al.(2022)Wang, Jansen, C{\^o}t{\'e}, and Ammanabrolu}]{wang2022scienceworld}
Ruoyao Wang, Peter Jansen, Marc-Alexandre C{\^o}t{\'e}, and Prithviraj Ammanabrolu. 2022.
\newblock Scienceworld: Is your agent smarter than a 5th grader?
\newblock In \emph{Proceedings of the 2022 Conference on Empirical Methods in Natural Language Processing}, pages 11279--11298.

\bibitem[{Wang et~al.(2023{\natexlab{a}})Wang, Dong, Cheng, Liu, Yan, Gao, and Wei}]{wang2023augmenting}
Weizhi Wang, Li~Dong, Hao Cheng, Xiaodong Liu, Xifeng Yan, Jianfeng Gao, and Furu Wei. 2023{\natexlab{a}}.
\newblock \href {https://openreview.net/forum?id=BryMFPQ4L6} {Augmenting language models with long-term memory}.
\newblock In \emph{Thirty-seventh Conference on Neural Information Processing Systems}.

\bibitem[{Wang et~al.(2023{\natexlab{b}})Wang, Cai, Chen, Liu, Ma, Liang, and CraftJarvis}]{wang2023describe}
Zihao Wang, Shaofei Cai, Guanzhou Chen, Anji Liu, Xiaojian Ma, Yitao Liang, and Team CraftJarvis. 2023{\natexlab{b}}.
\newblock Describe, explain, plan and select: interactive planning with large language models enables open-world multi-task agents.
\newblock In \emph{Proceedings of the 37th International Conference on Neural Information Processing Systems}, pages 34153--34189.

\bibitem[{Wang et~al.(2024{\natexlab{c}})Wang, Cai, Liu, Ma, and Liang}]{wang2024jarvis}
Zihao Wang, Shaofei Cai, Anji Liu, Xiaojian Ma, and Yitao Liang. 2024{\natexlab{c}}.
\newblock Jarvis-1: Open-world multi-task agents with memory-augmented multimodal language models.
\newblock In \emph{Second Agent Learning in Open-Endedness Workshop}.

\bibitem[{Wei et~al.(2022)Wei, Wang, Schuurmans, Bosma, Xia, Chi, Le, Zhou et~al.}]{wei2022chain}
Jason Wei, Xuezhi Wang, Dale Schuurmans, Maarten Bosma, Fei Xia, Ed~Chi, Quoc~V Le, Denny Zhou, et~al. 2022.
\newblock Chain-of-thought prompting elicits reasoning in large language models.
\newblock \emph{Advances in neural information processing systems}, 35:24824--24837.

\bibitem[{Wu et~al.(2024)Wu, Tang, Mitchell, and Li}]{wu2024smartplay}
Yue Wu, Xuan Tang, Tom Mitchell, and Yuanzhi Li. 2024.
\newblock Smartplay: A benchmark for llms as intelligent agents.
\newblock In \emph{The Twelfth International Conference on Learning Representations}.

\bibitem[{Xie et~al.(2024)Xie, Zhang, Chen, Li, Zhao, Cao, Hua, Cheng, Shin, Lei et~al.}]{xie2024osworld}
Tianbao Xie, Danyang Zhang, Jixuan Chen, Xiaochuan Li, Siheng Zhao, Ruisheng Cao, Toh~Jing Hua, Zhoujun Cheng, Dongchan Shin, Fangyu Lei, et~al. 2024.
\newblock Osworld: Benchmarking multimodal agents for open-ended tasks in real computer environments.
\newblock \emph{arXiv preprint arXiv:2404.07972}.

\bibitem[{Xu et~al.(2023)Xu, Wang, Li, Luo, Wang, Liu, and Liu}]{xu2023exploring}
Yuzhuang Xu, Shuo Wang, Peng Li, Fuwen Luo, Xiaolong Wang, Weidong Liu, and Yang Liu. 2023.
\newblock Exploring large language models for communication games: An empirical study on werewolf.
\newblock \emph{arXiv preprint arXiv:2309.04658}.

\bibitem[{Yang et~al.(2025)Yang, Chen, Zhang, Zhao, Qian, Wang, Wang, Koripella, Movahedi, Li et~al.}]{yang2025embodiedbench}
Rui Yang, Hanyang Chen, Junyu Zhang, Mark Zhao, Cheng Qian, Kangrui Wang, Qineng Wang, Teja~Venkat Koripella, Marziyeh Movahedi, Manling Li, et~al. 2025.
\newblock Embodiedbench: Comprehensive benchmarking multi-modal large language models for vision-driven embodied agents.
\newblock \emph{arXiv preprint arXiv:2502.09560}.

\bibitem[{Yao et~al.(2022)Yao, Chen, Yang, and Narasimhan}]{yao2022webshop}
Shunyu Yao, Howard Chen, John Yang, and Karthik Narasimhan. 2022.
\newblock Webshop: Towards scalable real-world web interaction with grounded language agents.
\newblock \emph{Advances in Neural Information Processing Systems}, 35:20744--20757.

\bibitem[{Yao et~al.(2024)Yao, Yu, Zhao, Shafran, Griffiths, Cao, and Narasimhan}]{yao2024tree}
Shunyu Yao, Dian Yu, Jeffrey Zhao, Izhak Shafran, Tom Griffiths, Yuan Cao, and Karthik Narasimhan. 2024.
\newblock Tree of thoughts: Deliberate problem solving with large language models.
\newblock \emph{Advances in Neural Information Processing Systems}, 36.

\bibitem[{Yao et~al.(2023)Yao, Zhao, Yu, Du, Shafran, Narasimhan, and Cao}]{yao2023react}
Shunyu Yao, Jeffrey Zhao, Dian Yu, Nan Du, Izhak Shafran, Karthik~R Narasimhan, and Yuan Cao. 2023.
\newblock React: Synergizing reasoning and acting in language models.
\newblock In \emph{The Eleventh International Conference on Learning Representations}.

\bibitem[{Young et~al.(2024)Young, Chen, Li, Huang, Zhang, Zhang, Li, Zhu, Chen, Chang et~al.}]{young2024yi}
Alex Young, Bei Chen, Chao Li, Chengen Huang, Ge~Zhang, Guanwei Zhang, Heng Li, Jiangcheng Zhu, Jianqun Chen, Jing Chang, et~al. 2024.
\newblock Yi: Open foundation models by 01. ai.
\newblock \emph{arXiv preprint arXiv:2403.04652}.

\bibitem[{Yu et~al.(2024)Yu, Yao, Li, Deng, Cao, Chen, Suchow, Liu, Cui, Xu, Zhang, Subbalakshmi, Xiong, He, Huang, Li, and Xie}]{yu2024finconsynthesizedllmmultiagent}
Yangyang Yu, Zhiyuan Yao, Haohang Li, Zhiyang Deng, Yupeng Cao, Zhi Chen, Jordan~W. Suchow, Rong Liu, Zhenyu Cui, Zhaozhuo Xu, Denghui Zhang, Koduvayur Subbalakshmi, Guojun Xiong, Yueru He, Jimin Huang, Dong Li, and Qianqian Xie. 2024.
\newblock \href {https://arxiv.org/abs/2407.06567} {Fincon: A synthesized llm multi-agent system with conceptual verbal reinforcement for enhanced financial decision making}.
\newblock \emph{Preprint}, arXiv:2407.06567.

\bibitem[{Yuan et~al.(2018)Yuan, C{\^o}t{\'e}, Sordoni, Laroche, Combes, Hausknecht, and Trischler}]{yuan2018counting}
Xingdi Yuan, Marc-Alexandre C{\^o}t{\'e}, Alessandro Sordoni, Romain Laroche, Remi Tachet~des Combes, Matthew Hausknecht, and Adam Trischler. 2018.
\newblock Counting to explore and generalize in text-based games.
\newblock \emph{arXiv preprint arXiv:1806.11525}.

\bibitem[{Zhang et~al.(2023)Zhang, Cai, Fu, Yuan, and Lu}]{zhang2023creative}
Chi Zhang, Penglin Cai, Yuhui Fu, Haoqi Yuan, and Zongqing Lu. 2023.
\newblock Creative agents: Empowering agents with imagination for creative tasks.
\newblock \emph{arXiv preprint arXiv:2312.02519}.

\bibitem[{Zhang et~al.(2024{\natexlab{a}})Zhang, Shen, Wu, Peng, Wang, Zhuang, and Lu}]{zhang2024self}
Wenqi Zhang, Yongliang Shen, Linjuan Wu, Qiuying Peng, Jun Wang, Yueting Zhuang, and Weiming Lu. 2024{\natexlab{a}}.
\newblock Self-contrast: Better reflection through inconsistent solving perspectives.
\newblock \emph{arXiv preprint arXiv:2401.02009}.

\bibitem[{Zhang et~al.(2024{\natexlab{b}})Zhang, Tang, Wu, Wang, Shen, Hou, Tan, Li, Zhuang, and Lu}]{zhang2024agent}
Wenqi Zhang, Ke~Tang, Hai Wu, Mengna Wang, Yongliang Shen, Guiyang Hou, Zeqi Tan, Peng Li, Yueting Zhuang, and Weiming Lu. 2024{\natexlab{b}}.
\newblock Agent-pro: Learning to evolve via policy-level reflection and optimization.
\newblock In \emph{ICLR 2024 Workshop on Large Language Model (LLM) Agents}.

\bibitem[{Zhang et~al.(2024{\natexlab{c}})Zhang, Li, Liu, Yu, Fung, Li, Li, and Ji}]{zhang2024knowledge}
Yuji Zhang, Sha Li, Jiateng Liu, Pengfei Yu, Yi~R Fung, Jing Li, Manling Li, and Heng Ji. 2024{\natexlab{c}}.
\newblock Knowledge overshadowing causes amalgamated hallucination in large language models.
\newblock \emph{arXiv preprint arXiv:2407.08039}.

\bibitem[{Zhao et~al.(2024)Zhao, Zhang, Li, Huang, Guo, Peng, Hao, Wen, Hu, Du, Guo, Li, and Chen}]{zhao2024assessingunderstandingcreativitylarge}
Yunpu Zhao, Rui Zhang, Wenyi Li, Di~Huang, Jiaming Guo, Shaohui Peng, Yifan Hao, Yuanbo Wen, Xing Hu, Zidong Du, Qi~Guo, Ling Li, and Yunji Chen. 2024.
\newblock \href {https://arxiv.org/abs/2401.12491} {Assessing and understanding creativity in large language models}.
\newblock \emph{Preprint}, arXiv:2401.12491.

\bibitem[{Zheng et~al.(2024)Zheng, Huang, Zhao, Zhong, and Wang}]{zheng2024towards}
Duo Zheng, Shijia Huang, Lin Zhao, Yiwu Zhong, and Liwei Wang. 2024.
\newblock Towards learning a generalist model for embodied navigation.
\newblock In \emph{Proceedings of the IEEE/CVF Conference on Computer Vision and Pattern Recognition}, pages 13624--13634.

\bibitem[{Zhong et~al.(2024)Zhong, Guo, Gao, Ye, and Wang}]{zhong2024memorybank}
Wanjun Zhong, Lianghong Guo, Qiqi Gao, He~Ye, and Yanlin Wang. 2024.
\newblock Memorybank: Enhancing large language models with long-term memory.
\newblock In \emph{Proceedings of the AAAI Conference on Artificial Intelligence}, volume~38, pages 19724--19731.

\bibitem[{Zhou et~al.(2024{\natexlab{a}})Zhou, Yan, Shlapentokh-Rothman, Wang, and Wang}]{zhou2024language}
Andy Zhou, Kai Yan, Michal Shlapentokh-Rothman, Haohan Wang, and Yu-Xiong Wang. 2024{\natexlab{a}}.
\newblock Language agent tree search unifies reasoning, acting, and planning in language models.
\newblock In \emph{Forty-first International Conference on Machine Learning}.

\bibitem[{Zhou et~al.(2023{\natexlab{a}})Zhou, Sch{\"a}rli, Hou, Wei, Scales, Wang, Schuurmans, Cui, Bousquet, Le et~al.}]{zhou2023least}
Denny Zhou, Nathanael Sch{\"a}rli, Le~Hou, Jason Wei, Nathan Scales, Xuezhi Wang, Dale Schuurmans, Claire Cui, Olivier Bousquet, Quoc~V Le, et~al. 2023{\natexlab{a}}.
\newblock Least-to-most prompting enables complex reasoning in large language models.
\newblock In \emph{The Eleventh International Conference on Learning Representations}.

\bibitem[{Zhou et~al.(2023{\natexlab{b}})Zhou, Xu, Zhu, Zhou, Lo, Sridhar, Cheng, Ou, Bisk, Fried, Alon, and Neubig}]{zhou2023webarena}
Shuyan Zhou, Frank Xu, Hao Zhu, Xuhui Zhou, Robert Lo, Abishek Sridhar, Xianyi Cheng, Tianyue Ou, Yonatan Bisk, Daniel Fried, Uri Alon, and Graham Neubig. 2023{\natexlab{b}}.
\newblock \href {https://openreview.net/forum?id=zlsj9akpaa} {Webarena: A realistic web environment for building autonomous agents}.
\newblock In \emph{NeurIPS 2023 Foundation Models for Decision Making Workshop}.

\bibitem[{Zhou et~al.(2024{\natexlab{b}})Zhou, Xu, Zhu, Zhou, Lo, Sridhar, Cheng, Ou, Bisk, Fried et~al.}]{zhou2024webarena}
Shuyan Zhou, Frank~F Xu, Hao Zhu, Xuhui Zhou, Robert Lo, Abishek Sridhar, Xianyi Cheng, Tianyue Ou, Yonatan Bisk, Daniel Fried, et~al. 2024{\natexlab{b}}.
\newblock Webarena: A realistic web environment for building autonomous agents.
\newblock In \emph{The Twelfth International Conference on Learning Representations}.

\bibitem[{Zhu et~al.(2025)Zhu, Du, Hong, Yang, Guo, Wang, Wang, Qian, Tang, Ji et~al.}]{zhu2025multiagentbench}
Kunlun Zhu, Hongyi Du, Zhaochen Hong, Xiaocheng Yang, Shuyi Guo, Zhe Wang, Zhenhailong Wang, Cheng Qian, Xiangru Tang, Heng Ji, et~al. 2025.
\newblock Multiagentbench: Evaluating the collaboration and competition of llm agents.
\newblock \emph{arXiv preprint arXiv:2503.01935}.

\bibitem[{Zhu et~al.(2023)Zhu, Chen, Tian, Tao, Su, Yang, Huang, Li, Lu, Wang et~al.}]{zhu2023ghost}
Xizhou Zhu, Yuntao Chen, Hao Tian, Chenxin Tao, Weijie Su, Chenyu Yang, Gao Huang, Bin Li, Lewei Lu, Xiaogang Wang, et~al. 2023.
\newblock Ghost in the minecraft: Generally capable agents for open-world environments via large language models with text-based knowledge and memory.
\newblock \emph{arXiv preprint arXiv:2305.17144}.

\end{thebibliography}

\clearpage
\appendix

\section*{Appendix}
\label{sec:appendix}

\section{Engine Design Details}
\label{apdx:engine_design_details}
The game engine involves scenes, tools, and items as three main components. We will introduce in detail each of them in the following.

\paragraph{Scene.} A scene typically includes a description, its connections to other scenes, and the tools and items it contains. A typical scene example in the game configuration looks like the following:\\
\makebox[\linewidth]{\rule{\linewidth}{0.4pt}}
\textit{Scene Example}
\begin{lstlisting}[basicstyle=\ttfamily\scriptsize, breaklines=true, breakindent=0em]
- name: hallway
  desc: You are in a hallway with a blocked path straight ahead, a locked cabinet on the left, and a corridor to the right.
  scene_relations:
    To the blocked path close-up: blocked path close-up
    To the cabinet close-up: cabinet close-up
  items:
    ...
  tools:
    ...
\end{lstlisting}
In this example, the name of this scene is ``hallway''. It leads to nearby scenes including ``blocked path close-up'' and ``cabinet close-up'', where the model could reach through action ``move(To the blocked path close-up)'' and ``move(To the cabinet close-up)''.

\paragraph{Tool.} Each tool has various states and a visibility status. In each state, a tool is either awaiting the application of another tool or ready to be applied to another tool or item. A typical tool example in the game configuration looks like the following:\\
\makebox[\linewidth]{\rule{\linewidth}{0.4pt}}
\textit{Tool Example}
\begin{lstlisting}[basicstyle=\ttfamily\scriptsize, breaklines=true, breakindent=0em]
- name: key
  visible: False
  states:
  - desc: A rusty silver key
    wait_for:
    - lubricant
  - desc: A silver key shining bright light, ready to use now
    apply_to:
    - safe
\end{lstlisting}
In this example, a rusty key (\textit{tool, state 1}) hidden in the chest (\textit{item}) won't be visible to the user until the chest is opened, and after applying lubricant (\textit{tool}), it may change to a non-rusty functional key (\textit{tool, state 2}) that could be applied to open a safe.

\paragraph{Item.} Item is an upgraded version of the tool, as each state may await multiple inputs or tools in order to trigger certain changes, including item and tool's visibility, state, etc. A typical item example in the game configuration looks like the following:\\
\makebox[\linewidth]{\rule{\linewidth}{0.4pt}}
\textit{Item Example}
\begin{lstlisting}[basicstyle=\ttfamily\scriptsize, breaklines=true, breakindent=0em]
- name: digital lock
  states:
  - desc: A digital lock linked to a card reader, power on now.
    transitions:
    - wait_for:
      - apply, card
      trigger:
      - change_state, 1
      reward: Authorization succeeds, you have to input a 4-digit password.
    - desc: A digital lock already authorized. The password panel awaits a 4-digit input.
      transitions:
      - wait_for:
        - input, 1672
        trigger:
        - change_state, item, cabinet door, 2
        - change_state, 2
        reward: The password is correct. A door opens somewhere ...
    - desc: A digital lock. You have already input the correct password.
\end{lstlisting}
For instance, a digital lock (\textit{item}) may await the application of card (\textit{tool}) for authorization and correct password input to trigger the closed cabinet door (\textit{item}, state 1) to open (\textit{item}, state 2).

\section{Data Annotation Details}
\label{apdx:data_annotation_details}
We recruited eight human annotators, all with prior Room Escape game experience (offline and online) and at least a bachelor's degree. To ensure a smooth annotation process, all annotators were U.S.-based students with computer science backgrounds. Each annotator received detailed guidelines to ensure objective annotations and was tasked with extracting game logic and object descriptions (scenes, items, and tools) based on the official guide. The foundational data logic will be released with the software, and all annotators consented to the data collection.

\section{Agent Design Details}
For both BaseAgent and EscapeAgent, we apply the same system prompt for its action-taking to ensure fairness:\\
\makebox[\linewidth]{\rule{\linewidth}{0.4pt}}
\textit{System Instruction}
\begin{lstlisting}[basicstyle=\ttfamily\scriptsize, breaklines=true, breakindent=0em]
You are in a Room Escape game. You should explore the scene and find out what to do next.
There are three types of interactives: items, which are the interactable things in the scene; tools, which are applicable tools in your bag; scenes, whcih are interactable scenes near your current position.

You can perform one of the following actions:
- click(<interactable item>): Click an <interactable item> to examine it or interact with it. For example, you can examine a door handle that is marked as interactable.
- apply(<applicable tool>, <interactable item>): Apply an <applicable tool> in your bag to an <interactable item>. For example, you can apply a key in your bag to an interactable locked door to open it.
- input(string, <interactable item>): Input a string (only digits and letters) to an <interactable item>. For example, you can input a string password to an interactable combination lock.
- move(<interactable scene>): Move to a nearby <interactable item> to further explore. For example, you can move to the living room to explore more interactable items there.
- craft(<applicable tool>, <applicable tool>): Use one <applicable tool> in bag to another <applicable tool> in bag to craft something new. For example, you can use a battery in bag to a controller in bag to craft a new charged controller.
For instance, some valid actions may be: click(microwave), apply(key, silver chest), craft(controller, battery), input(c79a1, combination lock), move(Go to operation room).
\end{lstlisting}
The system prompts explicit instructs on the agent's action space with examples. In the following, we present in this section more details on EscapeAgent design, including BaseAgent, Reflection, and Foresight modules.

\subsection{BaseAgent Details}
\label{apdx:base_agent}
At each step, the BaseAgent receives information from the game engine. This information typically includes an environment description, a list of interactable objects in the scene, and the tools available in the agent's bag. A typical environment description appears as follows:\\
\makebox[\linewidth]{\rule{\linewidth}{0.4pt}}
\begin{lstlisting}[basicstyle=\ttfamily\scriptsize, breaklines=true, breakindent=0em]
Scene Description:
You are in the scene 'underneath part of the van'. There is a stepladder on the right side. There is a license plate on the left side.
Here are the items you can see in this scene:
- On the left side, there is license plate space: The license plate is currently fixed to the van, with four screws on each corner
- On the right side, there is stepladder: The stepladder is unfolded, now you can reach the top of the van,

Possible Actions:
Here are all the items in the scene that you can perform 'click', 'apply' or 'input':
<interactable item> license plate space
Here are nearby scenes that you can perform 'move' to further explore:
<interactable scene> Back to the back of the van: It leads to back of the van

Tools in Bag:
Here are the tools in your bag. You can perform 'craft' to use two tools in your bag to craft a new one, or perfom 'apply' to apply one tool in your bag to an object in the scene:
<applicable tool> bunch of keys: a bunch of keys with a keychain and some rust on it
<applicable tool> rag: a rag soaked with engine oil"
\end{lstlisting}
The scene typically includes a general description, while each item within the scene is accompanied by a detailed description. Possible actions specify which items or aspects of the scene the agent can interact with, and tools in the bag indicate which tools are available for use.

This environment description will also be coupled with working memory of previous steps. Each step's memory contains the following fields:\\
\makebox[\linewidth]{\rule{\linewidth}{0.4pt}}
\begin{lstlisting}[basicstyle=\ttfamily\scriptsize, breaklines=true, breakindent=0em]
History: [Step ...]
Your position: <How you get to that position from the beginning scene> e.g. Living Room -> Outside Corridor
Your action: <The action taken> e.g. move(Explore the blocked path)
Response from the environment: <Feedback from game engine> e.g. Action executed successfully. Change to another scene: blocked path close-up.
\end{lstlisting}
Using this information, the BaseAgent is instructed to explicitly apply a Chain-of-Thought reasoning process to determine its next action. This action is then parsed and sent back to the game engine. The game engine updates its state based on the agent’s input and provides feedback to the agent. By default, this feedback indicates whether the action was successful or not. In Easy and Normal game settings, additional customized environment feedback is provided. However, in the Hard game setting, no extra information is given.

\subsection{Reflection Module Details}
\label{apdx:reflection_module}
The Reflection module is integrated as a downstream component after BaseAgent within the EscapeAgent design. This module is responsible for maintaining a task list that is updated solely based on the agent's current actions and the feedback received. Each task in the list generally includes the following fields:\\
\makebox[\linewidth]{\rule{\linewidth}{0.4pt}}
\begin{lstlisting}[basicstyle=\ttfamily\scriptsize, breaklines=true, breakindent=0em]
[Task Index <index>] Name: <brief task name>, Target Item: <item name>
- Task description: <description of the task>
Example Task:
[Task Index 1] Name: open the chest, Target Item: chest
- Task description: To open the chest wth a matched key, I have tried simple click, apply safe key but all failed.
\end{lstlisting}
The task index facilitates the identification of specific tasks during task list management operations. The target item specifies the item in the scene that the task is focused on, enhancing the agent’s sense of purpose when exploring and performing trials within the scene. The task description outlines a potential strategy for solving the task, including actions the agent has previously attempted but failed. The following system prompt is used to guide the model:\\
\makebox[\linewidth]{\rule{\linewidth}{0.4pt}}
\textit{System Instruction}
\begin{lstlisting}[basicstyle=\ttfamily\scriptsize, breaklines=true, breakindent=0em]
You are a helpful agent to reflect on your action and environment response, and then maintain a task list with solving suggestions.
The role of this task list is that there are some tasks you currently cannot solve with the tools at hand, but you think you may need to solve them later, so write them down with some suggestions and hints for your future reference.

After analyzing your current action and the response from the environment, you should give an action to maintain the task list: update ,new, delete or none.
- update(updated_feedback): The parameter should an updated feedback about what you newly tried but failed. The updated feedback should retain the original feedback, and add one new hindsight you got from current action.
- new(task_name, feedback): The first parameter should be a brief name of the new task you propose, the second parameter should be what you have to do (extract hint from environment response) to solve this task.
- delete(index): If you choose delete, then the first parameter should be the index of the task in the task list that you thought you have completed or is not useful anymore.
- none(): If you choose none, do not give any parameter, it indicates you believe you don't need to perform any action on the task list in the current step
For instance, valid task list maintaining action may be: update(The door has a keyhole and needs a key. I try apply a hammer but fails.), new(open the safe, I need a 4-digit password input to open it with a hint of sigma sign beside the safe.), delete(1), none().
\end{lstlisting}
Note that the operations ``update'' and ``delete'' can be implemented in a rule-based manner. An update is triggered when the model attempts an action on a specific item and fails, requiring revised feedback. A delete occurs when the model successfully performs an action that advances progress, necessitating the removal of the corresponding task, if it exists, from the task list.

\subsection{Foresight Module Details}
\label{apdx:foresight_module}
The Foresight module serves as an upstream component preceding the BaseAgent in the EscapeAgent design. This module is activated when a new tool is collected during the last action step or a new task is added during the previous reflection step.
When a new tool is collected, the agent is provided with the current task list and instructed to propose potential applications for the tool within the context of these tasks and their specific scenarios. Additionally, the agent is given a list of all existing tools in its bag and encouraged to creatively evaluate whether the new tool could be combined with others to craft something useful. The following system prompt is employed to guide the model:\\
\makebox[\linewidth]{\rule{\linewidth}{0.4pt}}
\textit{System Instruction}
\begin{lstlisting}[basicstyle=\ttfamily\scriptsize, breaklines=true, breakindent=0em]
You are in a Room Escape game. You have to use your creativity to figure out the use of the tool you have just collected.
There are generally two ways about how to use the tool:
1. Combine this tool with another one in your bag to craft a new tool. In this case, use acton 'craft(<collected tool>, <applicable tool>)', e.g. craft(controller, battery) indicates use a battery in your bag you already have to the controller you just collected to craft a charged controller.
2. Apply this tool to a target item in a task to try solve this task. In this case, use action 'apply(<collected tool>, Target Item in a task)', e.g. apply(key, locked cabinet) indicates apply the key you just collected to a locked cabinet to open it.

Here are some general hints that you may follow:
1. Please especially pay attention to the description of the task and the tool, try to find the connection between them to justify your action.
2. In your '- Thought: ...' part in response, you shuold explicitly think about whether there's item in bag for crafting, or task in the list for applying this tool. You should read and infer carefully from the tool descriptions and the task description, and evaluate one by one.
3. In your '- Actions: ...' part in response, you should give zero to multiple action calls. For each action, you should follow the format 'craft(<collected tool>, <applicable tool>)' or 'apply(<collected tool>, Target Item in a task)'. If it's a craft action, you should justify why crafting here makes sense. If it's an apply action, you should first give the task index corresponding to the target item, then justify why this tool may solve the task.
\end{lstlisting}

If a new task is created, the agent is provided with a list of all the tools currently in its bag. It is then instructed to reason creatively about which tools could be applied to address the newly created task. The following system prompt is used to guide the model:\\
\makebox[\linewidth]{\rule{\linewidth}{0.4pt}}
\textit{System Instruction}
\begin{lstlisting}[basicstyle=\ttfamily\scriptsize, breaklines=true, breakindent=0em]
You are in a Room Escape game. You have to use your creativity to figure out if you could use any tools you have now to solve a new task you have just discovered.
There are generally three ways to solve a task:
1. Click the target item to simply interact with it to solve the task. In this case, use action 'click(Target Item in current task)', e.g. click(microwave) indicates click the microwave to examine it and try solve the task.
2. Use the tool in your bag to apply to the target item in the task. In this case, use action 'apply(<applicable tool>, Target Item in current task)', e.g. apply(key, locked cabinet) indicates apply the key in your bag to a locked cabinet to open it.
3. Input a string to the target item in the task. In this case, use action 'input(<any string>, Target Item in current task)', e.g. input(2413, combination lock) indicates input a string password to the combination lock to solve the task.

Here are some general hints that you may follow:
1. Please especially pay attention to the description of the task about what might be needed. Please always first try simple click to interact if haven't done so. Examine the tool description and your memory pad, try to find the connection between them and what this task needs to justify your action.
2. In your '- Thought: ...' part in response, you should explicitly think about whether there's item to click, tool in bag for applying, or hint from memory pad and tools for string input. You should read and infer carefully from the task description, evaluate one by one.
3. In your '- Actions: ...' part in response, you should give zero to multiple action calls. For each action, you should follow the format 'click(Target Item in current task)', 'apply(<applicable tool>, Target Item in current task)', or 'input(<any string>, Target Item in current task)'. You shuold justify why this action may solve the task according to the task description, tool description, and memory pad hint.
\end{lstlisting}

Depending on whether the model proposes a valid action, the agent transitions into either the ``Free Explore'' or ``Try Action'' state. If multiple actions are proposed, the agent attempts them sequentially until a successful action is achieved. In cases where the task cannot be solved with the currently available tools, the task remains on the task list, and the newly acquired tool stays in the bag. For consistency, all action trials are included in the total count of action steps.

\begin{figure*}[!t]
    \centering
    \subfigure{\includegraphics[width=\linewidth]{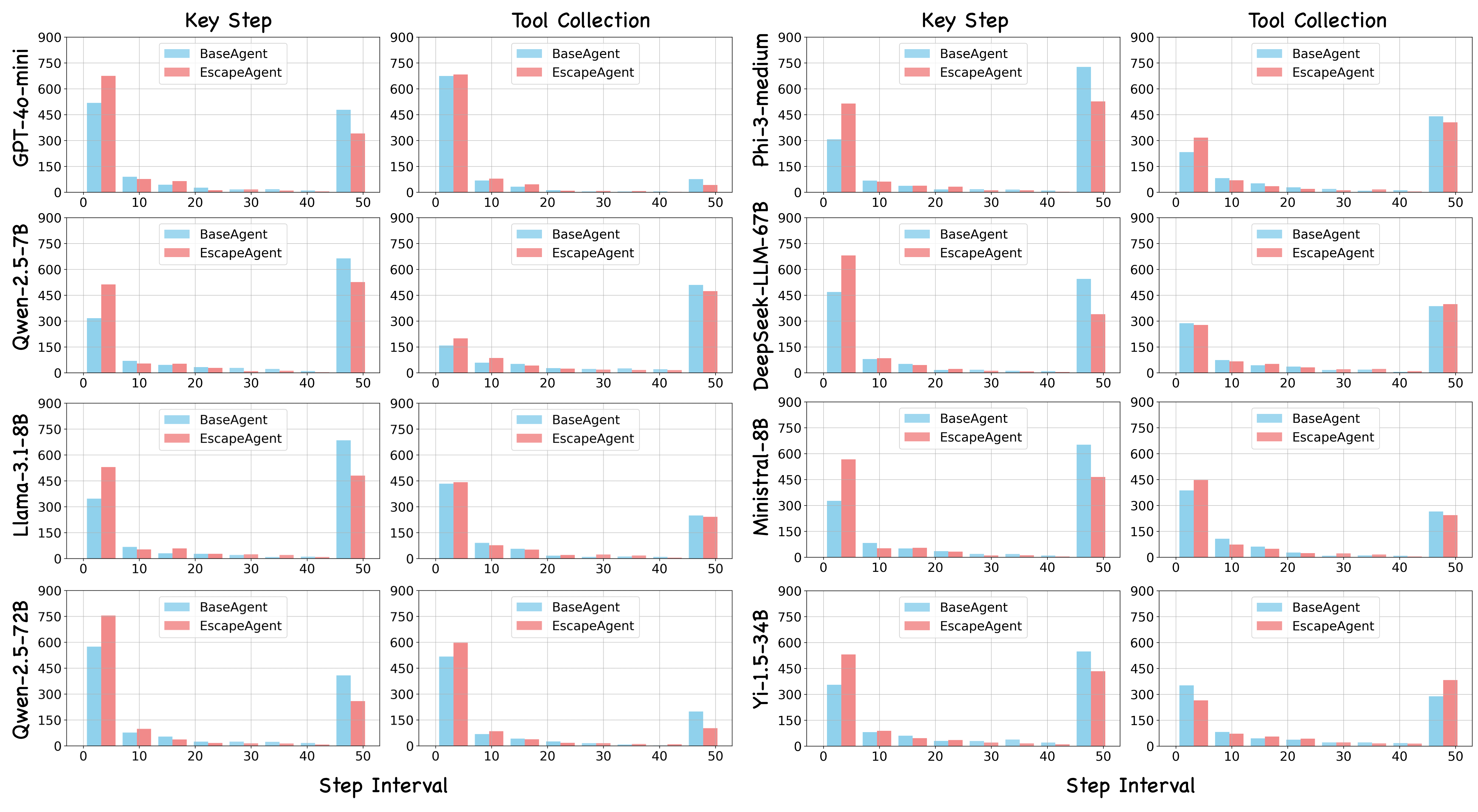}}
    \caption{More model's analysis on progress-making interval, extension of \Cref{fig:analysis_interval}.}
    \label{fig:analysis_interval_apdx}
\end{figure*}

\begin{figure*}[!t]
    \centering
    \subfigure{\includegraphics[width=\linewidth]{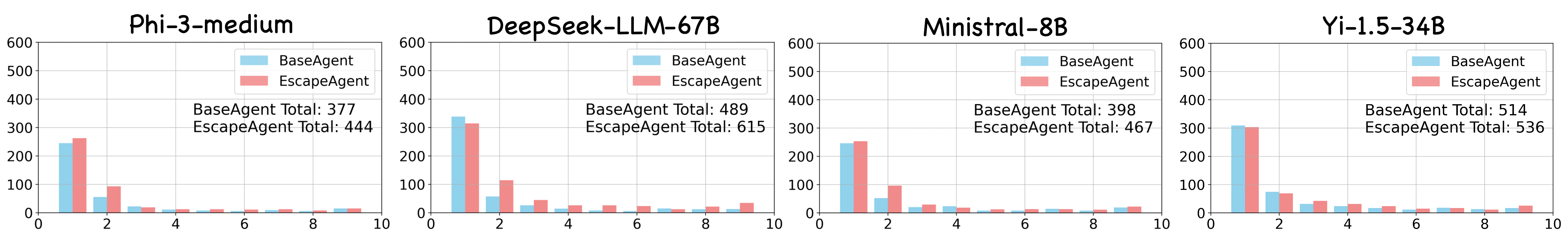}}
    \caption{More model's analysis on item trial times, extension of \Cref{fig:analysis_trialtime}.}
    \label{fig:analysis_trialtime_apdx}
\end{figure*}

\section{More Experiment Details}
\subsection{Help Setting}
\label{apdx:experiment_setting}
We provide help to the agent through explicit instructions, focusing on two aspects: i) identifying the next action location that could help the model make progress, and ii) specifying the action the model should take. These objectives are addressed by providing the instruction below to the model during the action-taking step:
\makebox[\linewidth]{\rule{\linewidth}{0.4pt}}
\textit{Action Instruction}
\begin{lstlisting}[basicstyle=\ttfamily\scriptsize, breaklines=true, breakindent=0em]
Since you're stuck, the system will provide you with a hint. You MUST follow the hint to complete next key step.
The next target location should be: <target position>.
Your next target action should be: <target action>.
You should go to the target position and perform the target action. If you are already at the target location, please directly perform the action.
\end{lstlisting}
This help will remain available until the model successfully performs the target action. All the helps on how to complete the game is provided through human annotation.

In each game, there may be cases where multiple actions can be taken in parallel since they do not interfere with one another. As a result, there is no fixed sequence among them. However, to simplify the help provided, the actions are linearized into a single action chain that ensures the agent can complete the game. Whenever the agent requires help, we always provide the first action in this chain that the model has not completed, even if the model may have already succeeded in performing some later actions in the chain.

\begin{figure*}[!t]
    \centering
    \subfigure{\includegraphics[width=\linewidth]{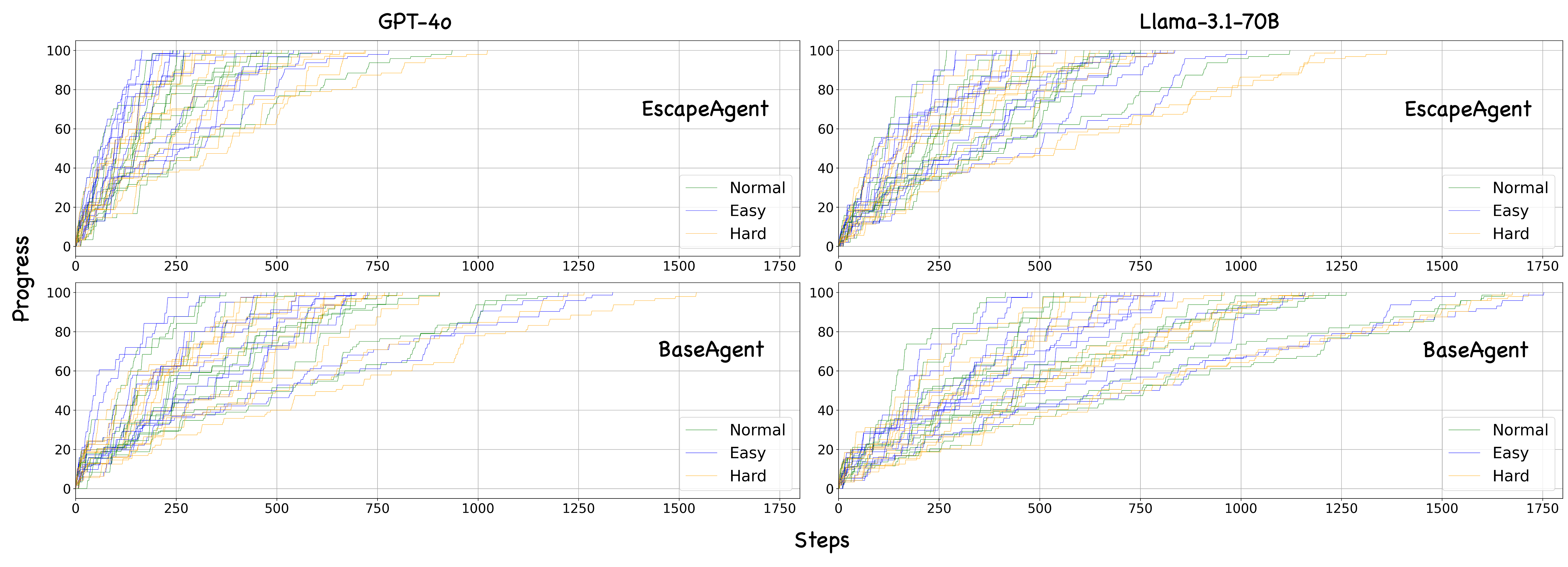}}
    \caption{An illustration of progress-making trend through all 36 game settings.}
    \label{fig:analysis_progress_all_apdx}
\end{figure*}

\subsection{Resource Setting}
\label{apdx:resource_setting}
For closed-source models, we utilize standard APIs for testing. Under the BaseAgent framework, the average number of API calls per game setting is approximately 800, resulting in a total cost of \$50–60 per model test. While the EscapeAgent framework introduces two additional modules, these are not always triggered. Consequently, the total number of API calls increases to roughly 1.2 times that of BaseAgent, raising the cost to approximately \$60–80 per model test.

For open-source models, all benchmarks are conducted using the vLLM framework on 2 A100-80G GPUs. Inference time varies based on model size: smaller-scale models complete all 36 game settings in approximately 12 hours, while larger 70B-scale models require about twice as much time for benchmarking. The average tool-calling frequency, reflected in the Total Steps metric, is reported in \Cref{tab:benchmark_results} for BaseAgent and \Cref{tab:creative_results} for EscapeAgent. These metrics vary significantly depending on the specific open-source models being tested.

\begin{figure}[!t]
    \centering
    \subfigure{\includegraphics[width=\linewidth]{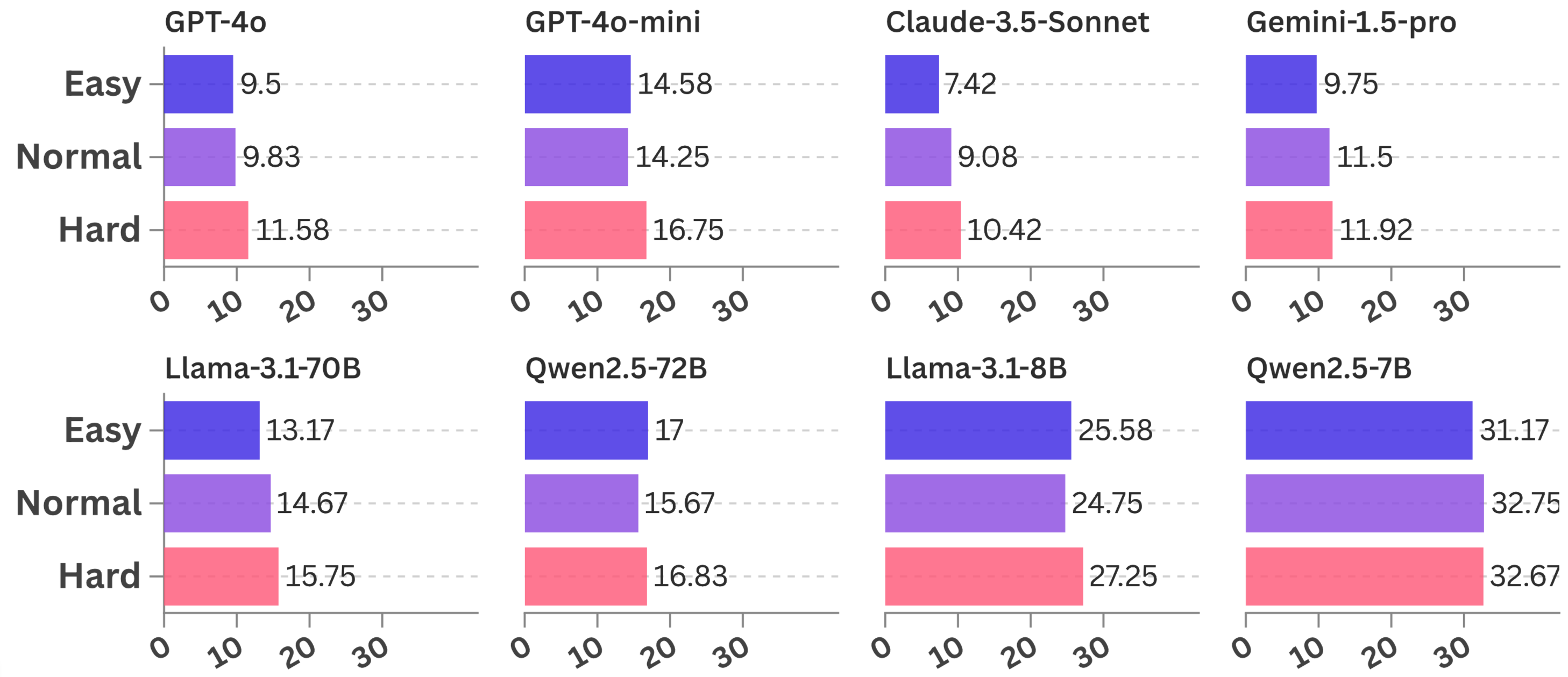}}
    \caption{Ablation of difficulty through Hints Used.}
    \label{fig:difficulty_hint}
\end{figure}

\begin{figure}[!t]
    \centering
    \subfigure{\includegraphics[width=\linewidth]{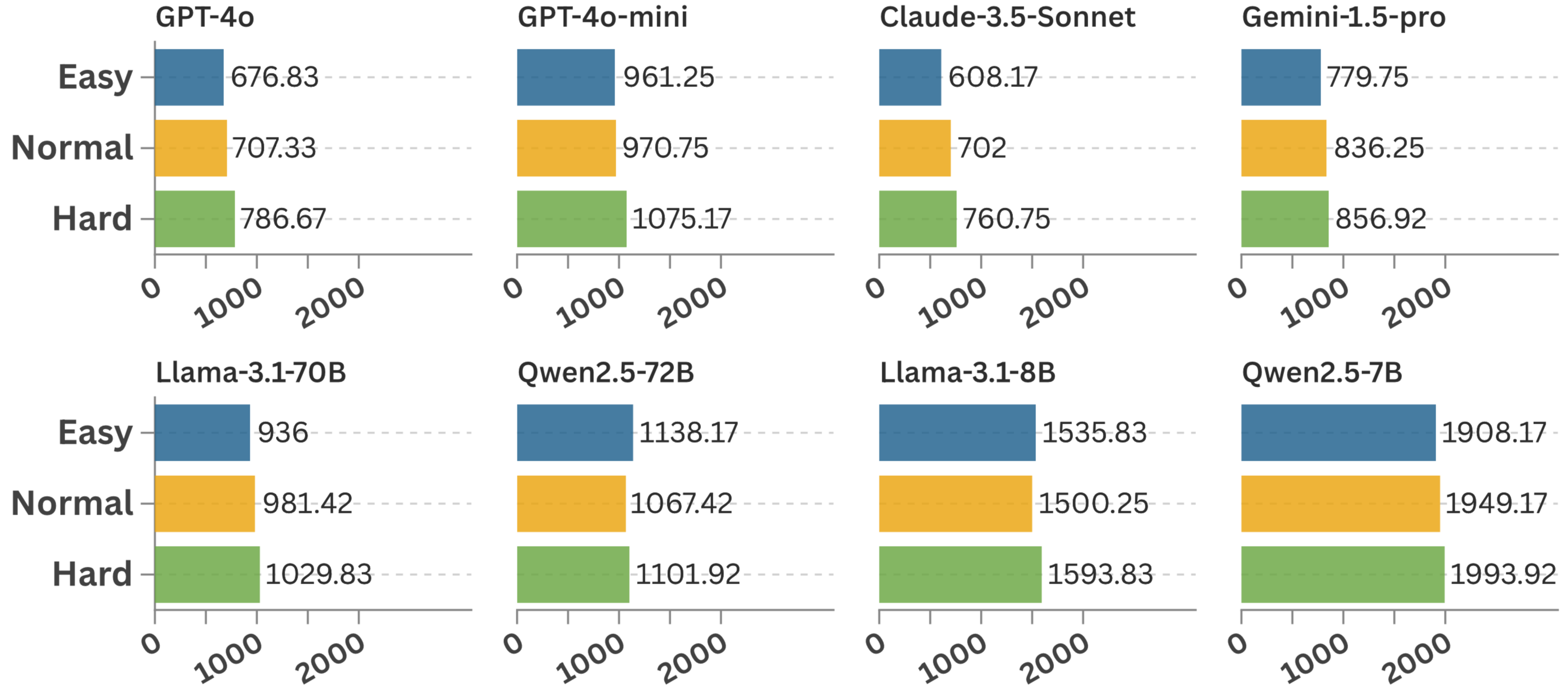}}
    \caption{Ablation of difficulty through Total Steps.}
    \label{fig:difficulty_step}
\end{figure}

\subsection{Ablation Study}
\label{apdx:ablation}
We perform an ablation study on the total steps and used hints in \Cref{fig:difficulty_hint} and \Cref{fig:difficulty_step}. Generally, harder game settings require more steps and hints for an agent to solve. Since our difficulty setting depends solely on the granularity and usefulness of descriptions and feedback (see \Cref{tab:difficulty_division}), our results demonstrate that the way the environment is presented can impact difficulty, even when the core game logic remains unchanged.

\begin{figure}[!t]
    \centering
    \subfigure{\includegraphics[width=\linewidth]{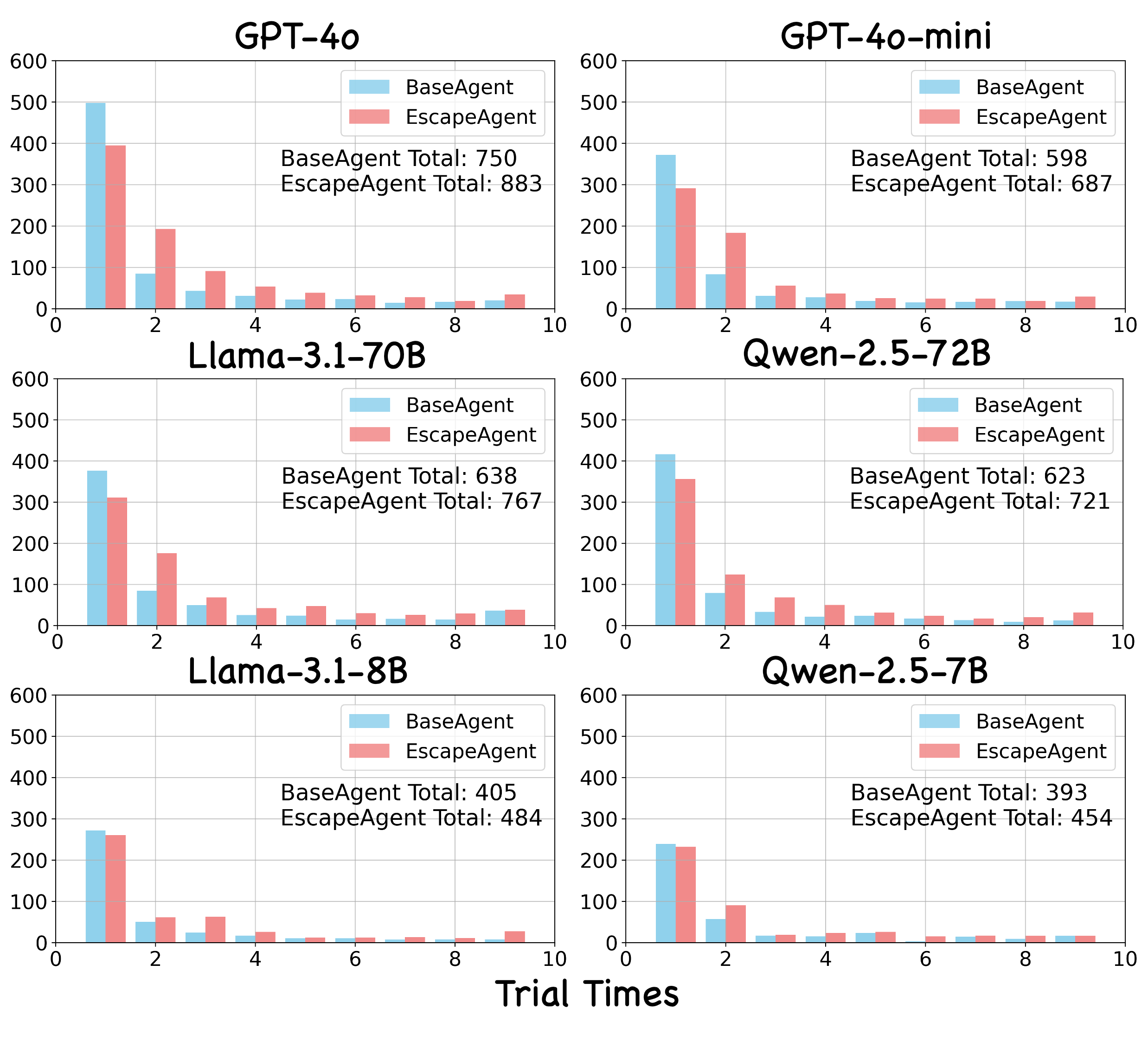}}
    \caption{A comparison of action trial times distribution for a specific item before success.}
    \label{fig:analysis_trialtime}
\end{figure}

\subsection{Further Analysis}
\label{apdx:trial}
We further analyze the valid actions agents attempt on different items in scenes before successfully operating them in \Cref{fig:analysis_trialtime}. While the BaseAgent exhibits a higher first-attempt success rate, EscapeAgent achieves greater effectiveness by making multiple attempts, leading to a significantly higher overall success rate within 10 trials. This difference can be attributed to EscapeAgent's strategy of proposing multiple viable actions simultaneously. Although this increases the likelihood of eventually succeeding by trying more actions, it does not prioritize the most-likely-to-succeed action first. As a result, the BaseAgent appears more efficient on its initial trial, but its superficial approach, as highlighted in \Cref{tab:error_analysis}, limits its overall performance. The EscapeAgent's design effectively addresses this limitation by leveraging a more exploratory approach, which proves advantageous in complex scenarios requiring creativity. 

\subsection{More Study Results}
\label{apdx:more_results}
In \Cref{fig:analysis_interval_apdx}, we present additional results on the action step intervals with respect to two types of progress-making ways, achieving Key Step or Tool Collection. It can be observed that EscapeAgent consistently requires fewer steps to perform the next bottleneck Key Step. In contrast, for Tool Collection, the difference between the two bars is less pronounced. Despite this, the findings still demonstrate the effectiveness of our design. Tool Collection typically occurs after the successful application or crafting of tools, meaning the reasoning challenge associated with it is significantly lower. Since EscapeAgent focuses primarily on creative reasoning, it is reasonable that it excels in identifying the next Key Action more efficiently.

In \Cref{fig:analysis_trialtime_apdx}, we present the results of action trial counts for a specific item across four additional models. We observe that EscapeAgent achieves a higher success rate within 10 trials, even though it does not always succeed on the first attempt. Furthermore, for smaller models like Phi-3 and Ministral, EscapeAgent occasionally outperforms BaseAgent even in terms of one-trial success rates. This highlights how our framework effectively lowers the barriers to creative reasoning, even for less capable language models.

\begin{figure*}[!t]
    \centering
    \subfigure{\includegraphics[width=\linewidth]{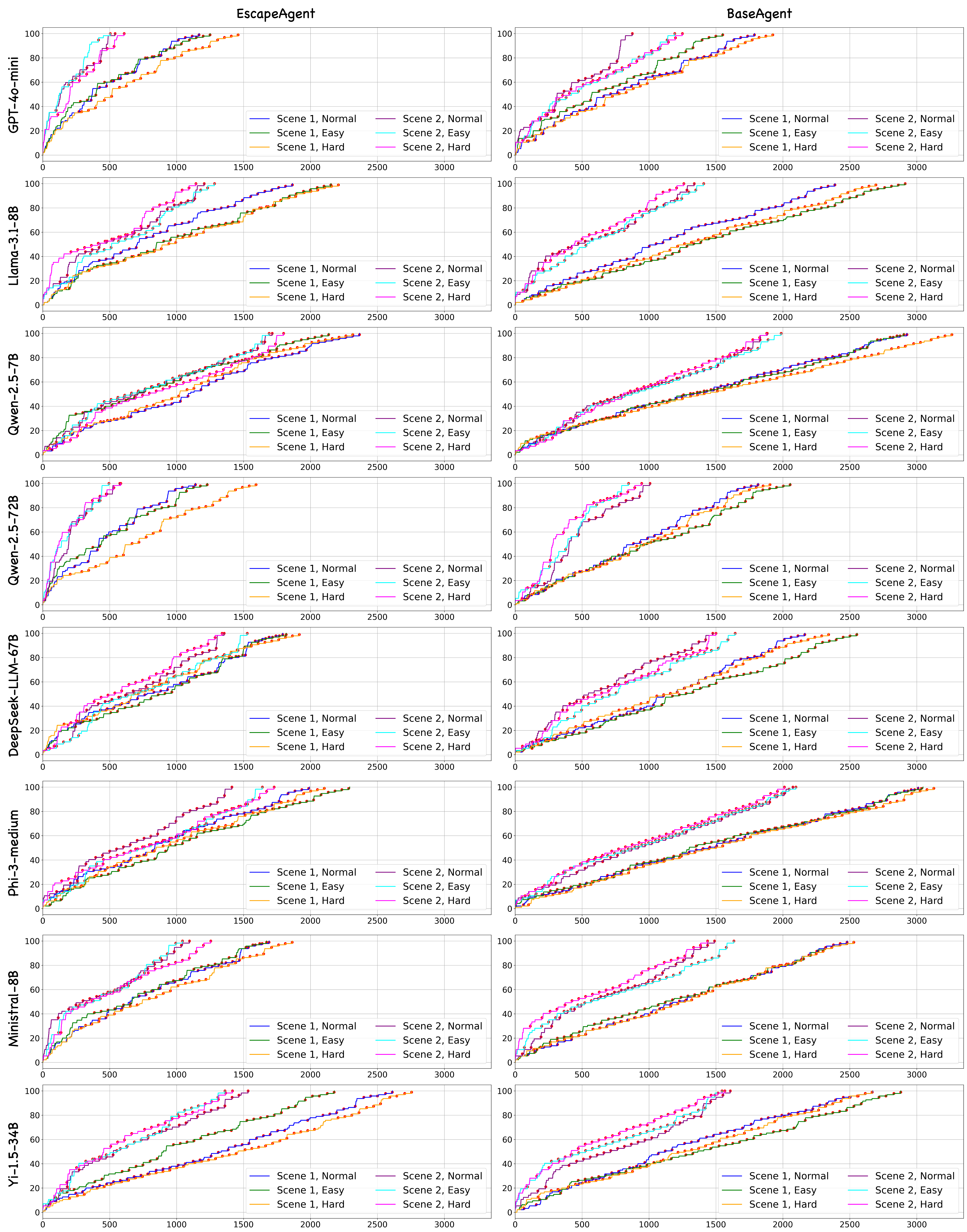}}
    \caption{More model's analysis on progress with respect to action steps, extension of \Cref{fig:analysis_progress}.}
    \label{fig:analysis_progress_apdx}
\end{figure*}

In \Cref{fig:analysis_progress_apdx}, we showcase eight additional pairs of progress-making maps for eight more models. These case studies illustrate two key points: i) There are significant disparities in creativity among models. For instance, models like GPT-4o-mini and Qwen-2.5-72B require only two-thirds of the total steps than others to achieve success, while smaller models such as Qwen-2.5-7B heavily rely on hints to make progress, even with the EscapeAgent framework. ii) When combining these results with \Cref{tab:creative_results}, we observe that EscapeAgent's performance improvement relative to BaseAgent is more pronounced for larger-scale models like GPT-4o and 70B-scale models. Conversely, while smaller models also benefit from reduced steps and hint usage, they still lag significantly behind in creative intelligence. This underscores the importance of enhancing a model's intrinsic reasoning abilities, as our method primarily mitigates the barriers to creative reasoning but does not fully address inherent limitations.

Lastly, \Cref{fig:analysis_progress_all_apdx} depicts the progress curves across all 36 game settings for GPT-4o and LLama-3.1-70B-Instruct. The trends reveal that BaseAgent's total step distribution spans a broader range compared to EscapeAgent, resulting in a relatively milder and less steep progression curve. These findings further confirm the effectiveness of EscapeAgent in facilitating more efficient progress.

\section{Human Performance Details}
\label{appendix:human_performance}

To better understand the ``Average Human'' baseline we present in the main table, we conducted a post-experiment survey with all participants, including:

\begin{itemize}[topsep=2pt, partopsep=-5pt, leftmargin=8pt, itemsep=-4.5pt]
    \item Prior offline Room Escape game experience? [Yes/No]
    \item Number of times played offline? [Integer]
    \item Prior online Room Escape experience (e.g., phone/PC)? [Yes/No]
    \item Self-assessed skill level? [Skilled / Might Be / Not Really / Not at All]
    \item Suggestions for improving EscapeBench
\end{itemize}

Results indicated that nearly all participants had prior experience, with an average of 5.6 offline sessions. However, only 20\% identified as skilled players; 40\% selected ``Might Be'' and another 40\% ``Not Really.'' Despite varying self-assessments, game logs showed no significant performance difference across groups—likely due to the trial-and-error nature of the tasks.

Although a large-scale user study is beyond our current scope, these findings suggest that our reported human performance offers a fair approximation of average human reasoning under novel conditions.

\section{Scalability of Benchmark Creation}
\label{apdx:bench_scalability}

To ensure high-quality annotations, we employ human annotators supported by an automated checker and manual cross-validation. While we experimented with GPT-4o for annotation, it faced two major issues: (i) frequent omission of interactable objects, and (ii) inability to balance difficulty across scenarios. These limitations currently require human intervention to maintain benchmark quality.

Specifically, we explored using GPT-4o to generate item-based scenarios. While the model can produce coherent logical sequences, it lacks the creativity and complexity required by our benchmark. For example:

$\bullet$ \noindent \textbf{Example 1:} Use a magnifying glass to examine an ancient scroll, revealing hidden formulas.  

$\bullet$ \noindent \textbf{Example 2:} Clean a rusty key with water, then scrape it with a coin to unlock a cabinet (partially creative).

Each AI-generated scenario yields about 12 key steps, whereas our benchmark includes over 1200 steps across 36 finely tiered scenarios. Therefore, human expertise remains essential for crafting creatively challenging and balanced content. While full automation falls short, parts of the process (e.g., validation, formatting) can be scaled. Future model improvements may increase automation potential, but human oversight will remain critical.

\section{More Detailed Discussions}
We present a more detailed version of our results discussions and future research directions.

\paragraph{Theoretical Foundations for AI Creativity.}
Understanding the cognitive mechanisms behind human creativity is essential for designing AI systems that emulate or surpass human creative processes. Human creativity is a multifaceted phenomenon involving the generation of novel and valuable ideas, problem-solving, and adaptation to complex situations \cite{dainys2024human}. Neurologically, it arises from the interplay between stochastic neuronal noise and structured, learned information, driven by spontaneous brain activity fluctuations \cite{malach2024neuronal}. This contrasts with AI’s reliance on data patterns and algorithmic processes. Boden identifies three AI creativity mechanisms: combining familiar ideas in novel ways, exploring conceptual spaces, and enabling transformative innovations \cite{boden1998creativity}. Our research shows that even advanced models like GPT-4o struggle with implicit goal identification and creative problem-solving, often requiring extensive prompting. The EscapeAgent framework significantly reduces this dependency and improves task-solving efficiency, indicating that these modules effectively overcome barriers to creative reasoning. However, the results also highlight the importance of the core model’s capabilities, as larger models like GPT-4o benefit more from the framework than smaller models. This suggests further research is needed to address current models' limitations in generating truly novel ideas. Interdisciplinary approaches integrating psychology, neuroscience, and model architecture can advance agent creativity further.

\paragraph{Multimodal Integration.}
Expanding the escape room environment to include multimodal data, such as visual and voice cues, offers a promising avenue for enhancing both the agent's performance and the realism of the scenarios. While integrating vision-language models could enable agents to interpret visual clues more naturally, such an extension would also require robust visual understanding and reasoning capabilities to handle the complexity of tasks effectively. Additionally, incorporating multimodal interactions presents opportunities to study how agents synthesize information across modalities, such as correlating visual patterns with instructions or adapting strategies based on dynamic, multimodal feedback. Future work should explore how to seamlessly integrate and harmonize these diverse data types into the benchmark, pushing the boundaries of agents' ability to process, reason about, and act on streams of information.

\paragraph{Step RL for Creative Reasoning.}
Introducing reinforcement learning into the EscapeBench task is expected to enhance the accuracy and efficiency of the agent’s exploration in long-chain tasks. This implies that under the guidance of rewards and penalties, the agent can explore in the correct direction more quickly. Compared with existing end-to-end reinforcement learning schemes, which rely on the final completion of the escape room task as the ultimate reward, introducing step rewards—providing immediate feedback for each step of the model’s operation—could potentially accelerate convergence and foster creative advancements in the model. Specifically, the discrete steps in EscapeBench can be organized either as a continuous, logically coherent progression or as strategic, abrupt jumps reflecting non-linear reasoning. Step Rewards can also further draw on the idea of task decomposition to be structured hierarchically. By constructing such a framework, Hierarchical Reinforcement Learning could manage super-long reasoning chains, decompose complex tasks into manageable subtasks, and enable a more fine-grained exploration and evaluation of AI’s creative capabilities.

\end{document}